\documentclass{article}

\usepackage[main,final]{neurips_2026}

\usepackage[utf8]{inputenc} %
\usepackage[T1]{fontenc}    %
\usepackage[dvipsnames]{xcolor}         %
\usepackage{hyperref}       %
\usepackage{url}            %
\usepackage{booktabs}       %
\usepackage{amsfonts}       %
\usepackage{nicefrac}       %
\usepackage{microtype}      %
\usepackage{mathtools}
\usepackage{amsthm}
\usepackage{amsmath}
\usepackage{amsfonts}
\usepackage{amssymb}
\usepackage{enumitem} %
\usepackage[capitalise, nameinlink]{cleveref}
\usepackage{colortbl}
\usepackage{geometry}
\usepackage{subcaption}
\usepackage{graphicx}
\usepackage{multirow}
\usepackage{wrapfig}
\usepackage{needspace}
\usepackage{tabularx}
\usepackage{afterpage}
\usepackage{tikz}

\newenvironment{itemizesquish}{\begin{itemize}[leftmargin=*,noitemsep,topsep=0pt]}{\end{itemize}}

\newcommand\numberthis{\addtocounter{equation}{1}\tag{\theequation}}
\definecolor{sprow}{gray}{0.92}

\newcommand{\patchSizePanel}[1]{%
\begin{tikzpicture}
    \node[anchor=south west, inner sep=0] (img) at (0,0) {
        \includegraphics[width=\linewidth]{#1}
    };
    \begin{scope}[x={(img.south east)}, y={(img.north west)}]
        \fill[white] (0.25,0.000) rectangle (0.75,0.055);
        \node[
            anchor=south,
            text=black!80,
            font=\fontfamily{phv}\fontsize{4.3}{5.6}\selectfont
        ] at (0.530,-0.015) {Avg. Bytes per Patch};
    \end{scope}
\end{tikzpicture}%
}

\title{Scratchpad Patching: Decoupling Compute from Patch Size in Byte-Level Language Models}

\author{
  {\bf Lin Zheng}\textsuperscript{1,2}\thanks{Work done during an internship at Google DeepMind.} \quad
  {\bf Vasilisa Bashlovkina}\textsuperscript{1} \quad
  {\bf Timothy Dozat}\textsuperscript{1} \\
  {\bf Dan Garrette}\textsuperscript{1} \quad
  {\bf Laura Rimell}\textsuperscript{1} \quad
  {\bf Joshua Maynez}\textsuperscript{1} \\
  \normalfont\textsuperscript{1}Google DeepMind \quad \textsuperscript{2}The University of Hong Kong \\
  \texttt{linzhengs@google.com}
}

\hypersetup{
  hidelinks,
  pdftitle={Scratchpad Patching: Decoupling Compute from Patch Size in Byte-Level Language Models},
  pdfauthor={Lin Zheng, Vasilisa Bashlovkina, Timothy Dozat, Dan Garrette, Laura Rimell, Joshua Maynez}
}

\begin{document}

\maketitle

\begin{abstract}
Tokenizer-free language models eliminate the tokenizer step of the language modeling pipeline by operating directly on bytes; patch-based variants further aggregate contiguous byte spans into patches for efficiency.
However, the average patch size chosen at the model design stage governs a tight trade-off: larger patches reduce compute and KV-cache footprint, but degrade modeling quality.
We trace this trade-off to \emph{patch lag}: until a patch is fully observed, byte predictions within it must rely on a stale representation from the previous patch to preserve causality; this lag widens as patches grow larger.
We introduce \emph{Scratchpad Patching} (SP), which inserts transient scratchpads inside each patch to aggregate the bytes seen so far and refresh patch-level context for subsequent predictions.
SP triggers scratchpads using next-byte prediction entropy, selectively allocating compute to information-dense regions and enabling post-hoc adjustment of inference-time compute.
Across experiments on natural language and code, SP improves model quality at the same patch size; for example, even at $16$ bytes per patch, SP-augmented models match or closely approach the byte-level baseline on downstream evaluations while using a $16\times$ smaller KV-cache over patches and requiring roughly $3$--$4\times$ less inference compute than the byte-level baseline.
\end{abstract}

\section{Introduction}
\label{intro}

Modern language models rely on tokenization \citep{sennrich2016bpe,kudo2018sentencepiece} to derive input representations and segment text into shorter token sequences. This handcrafted, non-end-to-end process introduces distinct drawbacks: the sequence shortening achieved by a fixed tokenizer is difficult to adapt or scale \citep{yu2025scone}, the model is sensitive to prompt formatting \citep{microsoft2023guidance,lundberg2023tokenhealing}, and glitch tokens can disrupt inference \citep{rumbelow2023solidgoldmagikarp,land2024fishing,yang2024problematictokens}. Recent research has therefore pivoted toward \emph{tokenizer-free} modeling---methods that operate directly on bytes without an externally defined subword vocabulary \citep{sutskever2011generating,graves2013generating,radford2017learning,chung2017hierarchical,hwang2017character,alrfou2019character,choe2019bridging,xue2022byt5,clark2022canine,wang2024mambabyte,zheng2025evabyte}. To mitigate the prohibitive cost of long byte sequences, \emph{patch-based} tokenizer-free models (\cref{background}) aggregate contiguous bytes into higher-level \emph{patches}, shortening the effective sequence length \citep{clark2022canine,nawrot2022hourglass,tay2022charformer,yu2023megabyte,nawrot2023dynamicpool,slagle2024spacebyte,ahia2024magnet,pagnoni2024blt,neitemeier2025hierarchical,owodunni2025flexitokens,videau2025aunets,hwang2025hnet,minixhofer2025bolmo}.

While promising, the standard approach to segmentation and formation of patch representations introduces a tight trade-off. Larger patch sizes yield fewer patches per input, improving computational efficiency and reducing KV-cache usage, but they also update patch-level context less frequently, forcing more byte predictions to be made from stale patch-level context. We call this staleness \emph{patch lag}. In a standard autoregressive patch-based model, only the final byte within each patch can use the completed representation of that patch, while every earlier byte must rely on the previous patch-level context to preserve causality (\cref{background:byte-lm}). As patches grow larger, this lag widens and makes modeling quality increasingly sensitive to patch size.

\begin{figure*}[t]
    \centering
    \includegraphics[width=\textwidth]{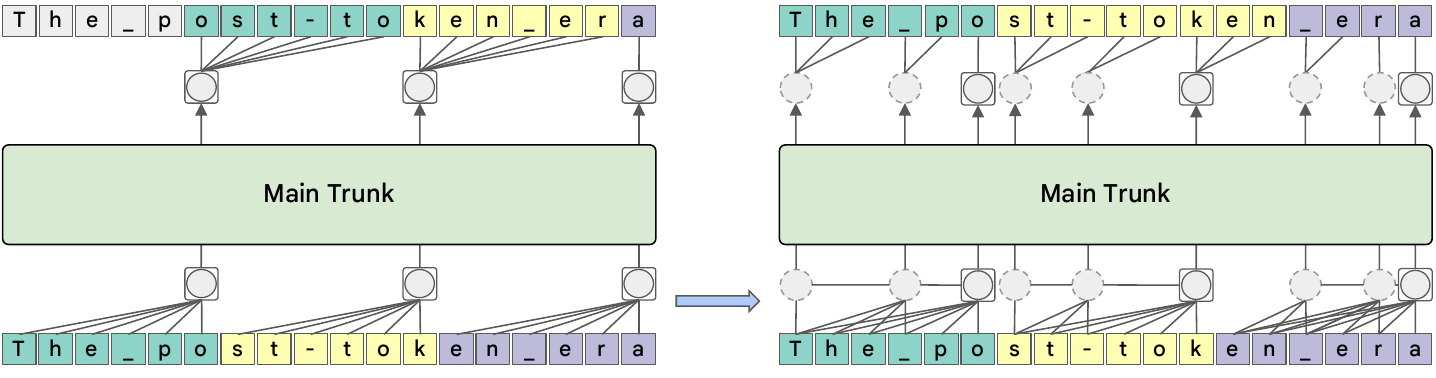}
    \caption{\textbf{Scratchpad Patching (SP).} \emph{Left:} a standard patch-based byte-level model runs the trunk $\mathcal{M}$ once per patch (see \cref{fig:compressive-arch} for the full architecture), leaving most byte predictions to rely on a stale representation from the previous patch. \emph{Right:} SP inserts transient scratchpads at selected byte positions, each aggregating the bytes seen so far within the patch and refreshing the trunk representation for subsequent predictions. This reduces patch lag while leaving the number of committed patch states unchanged. Other architectural components are omitted for clarity.}
    \label{fig:sp}
    \vspace{-5pt}
\end{figure*}

In this work, we introduce \emph{Scratchpad Patching} (SP), which decouples compute allocation from patch size to address patch lag (\cref{method}). Rather than committing a single representation only at each patch boundary, SP inserts transient \emph{scratchpads} at selected internal byte positions (\cref{fig:sp}). Each scratchpad aggregates the bytes seen so far within the patch and serves subsequent byte predictions until the next scratchpad or the committed patch representation is produced (\cref{method:main}). Because within-patch scratchpads are excluded from the persistent KV cache at inference, they leave the committed patch sequence length unchanged (\cref{method:impl}). Among several strategies we evaluate, triggering scratchpads via next-byte prediction entropy is most effective, selectively allocating compute to information-dense regions; the same machinery also enables post-hoc adjustment of inference-time compute without retraining.

SP is compatible with all patchifier families we test. Across experiments, SP improves the empirical frontier of quality versus patch size (\cref{experiments:main}): models can use larger patches and smaller KV caches without the usual quality penalty. With SP in place, different patching strategies from previous work cluster in performance-FLOPs space, indicating that the primary bottleneck may be insufficient compute rather than suboptimal boundary placement (\cref{experiments:compute-allocation}). Further analyses show that under FLOPs-matched comparisons, SP matches or improves non-SP baselines on three of the four patchifier families, suggesting that much of the gain comes from better-targeted rather than additional compute (\cref{analyses:flops-matched-performance}). Our contributions are as follows.
\begin{itemizesquish}
    \item We introduce \emph{Scratchpad Patching}, a mechanism that decouples compute from patch size to reduce \emph{patch lag}, which we characterize as a structural failure mode of patch-based models.
    \item We show that SP improves the empirical frontier of quality versus patch sequence length across downstream tasks; even at $16$ bytes per patch, SP models can match or closely approach the byte-level baseline while using a $16\times$ smaller trunk KV-cache over patches and requiring roughly $3$--$4\times$ less inference compute than the byte-level baseline.
    \item We find that with SP in place, the performance gap among patchifier families narrows substantially under comparable FLOPs budgets, and simple schemes such as fixed-size patching become competitive with complex boundary strategies.
\end{itemizesquish}

\section{Background}
\label{background}

\subsection{Tokenizer-based Language Modeling}
\label{background:token-based-lm}
Most modern language models operate on tokenized text \citep{bengio2003neural,devlin-etal-2019-bert,gpt3,openai2023gpt4,google2023gemini}. Given a raw text string, a tokenizer maps it to a discrete sequence of tokens $t_1, \dots, t_M$, where each token typically corresponds to a subword unit \citep{gage1994bpe,schuster2012wordpiece,wu2016googlenmt,sennrich2016bpe,kudo2018sentencepiece,kudo2018subwordregularization,dagan2024getting,liu2025superbpe}. The model is trained to maximize the log-likelihood $\sum_{i=1}^{M} \log p(t_i \mid t_{<i})$ of the observed token sequence.

Tokenization reduces the input sequence length and defines tokens as the \emph{atomic} prediction units of the model. While effective, this external preprocessing step couples the model to a fixed segmentation scheme and can introduce brittleness (\cref{intro}).

\subsection{Patch-based Byte-level Modeling}
\label{background:byte-lm}
These limitations have motivated tokenizer-free approaches that operate directly on bytes. In byte-level language modeling, the input becomes a UTF-8 byte sequence $b_1, \dots, b_N$, where each $b_i \in \{0, \dots, 255\}$.\footnote{In practice we expand the vocabulary beyond 256 to reserve IDs for sentinel tokens. In our experiments, the vocabulary size is set to $320$ with the last $64$ IDs reserved for sentinels such as \texttt{<bos>} and \texttt{<pad>}.} The model defines an autoregressive distribution $p(b_i \mid b_{<i})$ to enable end-to-end modeling without tokenization. Because byte sequences are substantially longer than token sequences, a recent line of work explores \emph{patch-based byte-level models}, which aggregate contiguous bytes into \emph{patches} and reduce the number of sequence elements processed by the main trunk.

\begin{wrapfigure}{r}{0.52\textwidth}
    \centering
    \includegraphics[width=0.5\textwidth]{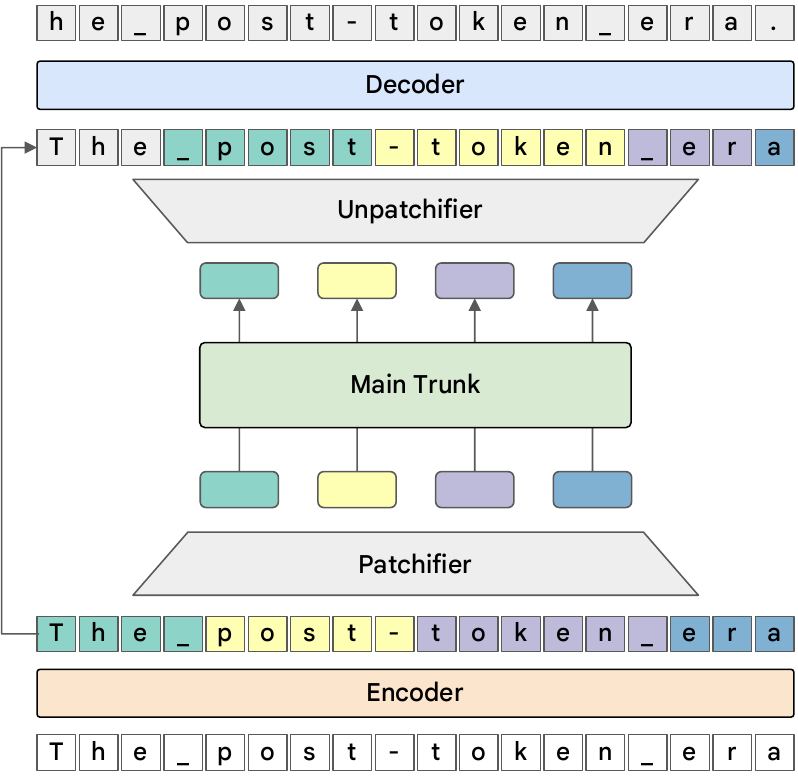}
    \caption{\textbf{Patch-based byte-level architecture.}}
    \label{fig:compressive-arch}
    \vspace{-10pt}
\end{wrapfigure}

\paragraph{Architecture.}
Most patch-based architectures share a design with five components (\cref{fig:compressive-arch}): an \emph{encoder}, a \emph{patchifier}, a \emph{main trunk}, an \emph{unpatchifier}, and a \emph{decoder}. The encoder $\mathcal{E}$, main trunk $\mathcal{M}$, and decoder $\mathcal{D}$ are all stacks of causal Transformer layers, while the patchifier and unpatchifier mediate between byte-level and patch-level representations.

The encoder $\mathcal{E}$ maps the byte sequence to contextual representations $x = \mathcal{E}(b)$. The patchifier $\mathcal{P}$ partitions the byte sequence into $L$ contiguous segments $[s_\ell, e_\ell]$ for each $\ell\in\{1,\dots,L\}$ and produces patch-level representations $z_\ell \coloneq \operatorname{Aggregate}\left(x_{s_\ell:e_\ell}\right)$ via local cross-attention, using the mean-pooled segment embedding as the query. Together with a \texttt{<bos>} sentinel $z_0$, these form the patch sequence $z_0, z_1, \dots, z_L$. The main trunk $\mathcal{M}$, which allocates the majority of model parameters and compute, processes the patch sequence as $\widetilde{z} = \mathcal{M}(z)$.

The unpatchifier $\mathcal{U}$ lifts patch-level trunk outputs back to byte positions and fuses them with encoder outputs $x$ via a residual connection \citep{hwang2025hnet}. Causality introduces an asymmetry: only the final byte of each patch ($n = e_\ell$) can condition on the \emph{current} patch's trunk output, while all earlier bytes must instead rely on the output of the \emph{previous} patch. We refer to the gap between a byte's prediction and the most recent patch-level representation available to it as \emph{patch lag}. In our backbone, it takes the following form\footnote{We omit linear projections that match trunk and encoder outputs to the decoder dimension; see \cref{app:impl_details}.}
\begin{align*}
    u_n = 
    \begin{cases}
        \widetilde{z}_{\ell-1} + x_n, &\text{ if } n \neq e_\ell,\\
        \widetilde{z}_{\ell} + x_n, &\text{ if } n = e_\ell.
    \end{cases}
    \numberthis \label{eqn:unpatchifier}
\end{align*}
Finally, the decoder $\mathcal{D}$ maps the resulting byte-level representations to next-byte prediction logits.

\paragraph{Patch Lag.}
Standard patch-based models treat each patch as an \emph{atomic} unit in the trunk. Consequently, trunk compute is governed primarily by the number of patches $L$, regardless of how many bytes or how much internal structure each patch represents. This tightly couples the capacity to patch size: as the average bytes per patch grow, patch lag widens, where non-final byte positions condition on an increasingly stale patch-level representation, resulting in the trade-off between shorter sequences and modeling quality. Our approach, introduced next, directly addresses this limitation.

\section{Scratchpad Patching}
\label{method}

Scratchpad Patching (SP) reduces patch lag without altering the patch sequence, decoupling compute allocation from patch size. Instead of mapping each patch to a single representation, SP introduces a sequence of \emph{scratchpad states} that progressively refine the patch representation by aggregating successively longer spans of bytes within the patch and passing each through the trunk. Because these states are used for computation but not persisted in the KV cache, each patch can undergo multiple internal refinement steps without increasing the inference-time trunk KV-cache footprint. \cref{fig:sp} provides intuition; we formalize scratchpad states and their interaction with patchification.

\subsection{Patchification with Scratchpads}
\label{method:main}

For each patch $\ell$ spanning byte positions $[s_\ell, e_\ell]$, SP associates each position $n$ with a binary indicator $p_n \in \{0,1\}$ specifying whether a scratchpad update fires at $n$. SP is agnostic to the choice of patchifier; if a position is both a patch boundary and a scratchpad trigger, patchification takes precedence and the scratchpad update is suppressed. These indicators induce a sequence of scratchpad states $z_\ell^{1}, \dots, z_\ell^{T_\ell}$ for patch $\ell$, where $T_\ell \coloneq \sum_{j=s_\ell}^{e_\ell} p_j$ counts the total updates and $T_\ell = 0$ recovers the standard patch-based model. $T_\ell$ may vary across patches, allowing the model to adaptively allocate more compute to longer or more information-dense patches. We reserve $z_\ell$ for the \emph{committed} patch representation $z_\ell \coloneq \operatorname{Aggregate}(x_{s_\ell:e_\ell})$, and $z_\ell^t$ for the \emph{transient} $t$-th scratchpad.

For any position $s_\ell \leq n \leq e_\ell$, let $t = \sum_{j=s_\ell}^{n} p_j$ index the scratchpad fired so far in patch $\ell$. When $p_n = 1$, we form $z_\ell^{t} = \operatorname{Aggregate}(x_{s_\ell:n})$ over this prefix and pass it through $\mathcal{M}$ in the same way as a regular patch state, yielding $\widetilde{z}_\ell^{t}$, then broadcast to byte positions for the decoder $\mathcal{D}$. Adopting the convention $\widetilde{z}_\ell^{0} \coloneq \widetilde{z}_{\ell-1}$ before any scratchpad fires within the current patch, \cref{eqn:unpatchifier} becomes
\begin{align*}
    u_n =
    \begin{cases}
        \widetilde{z}_{\ell}^{t} + x_n, & \text{if } n \neq e_\ell,\\
        \widetilde{z}_{\ell} + x_n, & \text{if } n = e_\ell.
    \end{cases}
    \numberthis \label{eqn:sp:unpatchifier}
\end{align*}
The essence of SP is replacing $\widetilde{z}_{\ell-1}$ in \cref{eqn:unpatchifier} with $\widetilde{z}_{\ell}^{t}$: after the first update, each non-final byte conditions on a fresh scratchpad state from the \emph{current} patch, rather than the stale representation from the previous patch. Patch lag is reduced from one full patch to the gap to the latest scratchpad.

\paragraph{Selective Scratchpad Updating.}

\begin{figure*}[t]
    \centering
    \includegraphics[width=\textwidth]{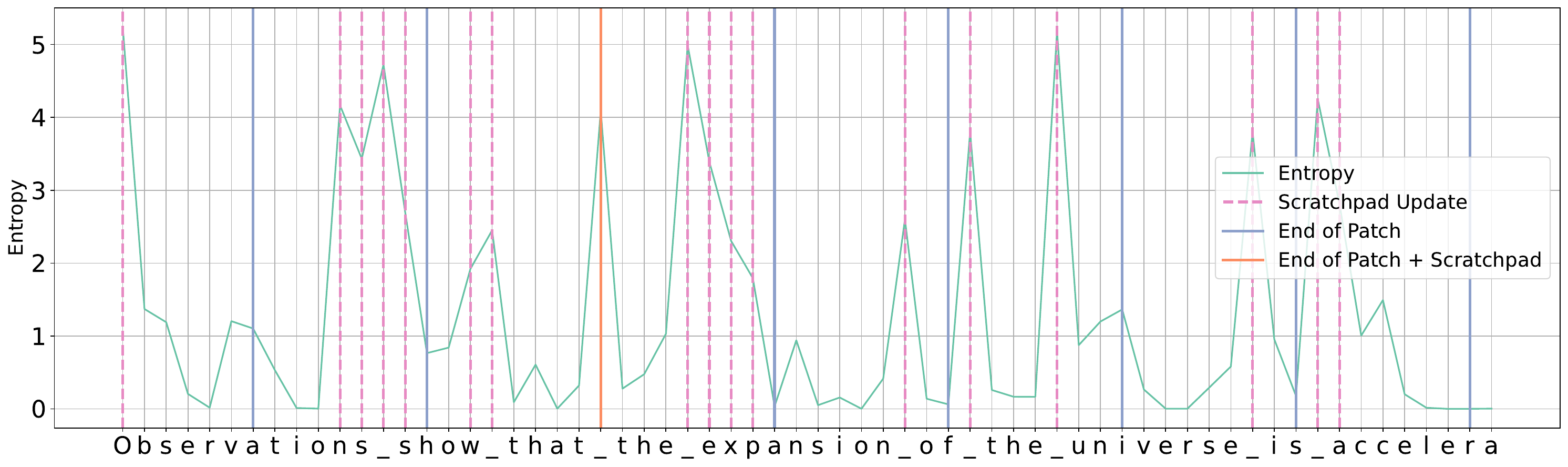} 
    \caption{\textbf{Scratchpad Patching dynamics on fixed-size patching ($p\!=\!8$).} Patch boundaries (\textbf{solid blue}) are regular by construction, while scratchpad updates (\textbf{dashed pink}) are triggered adaptively whenever the encoder's next-byte entropy (\textbf{green}) exceeds threshold $\tau_\text{SP} = 1.5$. When a scratchpad-trigger coincides with a patch boundary, patchification takes precedence (\textbf{solid orange}).}
    \label{fig:patching_demo:fixed8:text}
    \vspace{-8pt}
\end{figure*}

A simple instantiation of SP applies a scratchpad update at every byte position. This minimizes patch lag but incurs compute comparable to a vanilla byte-level model, negating the compute savings of patchification. Empirically, such dense updates also yield diminishing returns over selective updating (\cref{analyses:sp_impl_ablation}). For adaptive, content-aware compute allocation, we instead parameterize the trigger using next-byte prediction entropy, derived from a language modeling (LM) head applied to the encoder outputs $x$. Specifically, a scratchpad update is issued whenever the encoder's prediction entropy $H_n \coloneq -\sum_{b \in \mathcal{V}}p(b \mid x_{\leq n})\,\log p(b \mid x_{\leq n})$ exceeds a predefined threshold: $p_n \coloneq \mathbf{1}_{[H_n > \tau_\text{SP}]}$. \cref{fig:patching_demo:fixed8:text} illustrates this on a sample sequence with fixed-size patching: scratchpad updates fire at positions of elevated next-byte entropy, while patch boundaries remain on a regular fixed-size grid. We ablate updating strategies in \cref{analyses:sp_impl_ablation} and provide additional qualitative case studies in \cref{app:p_and_sp_case_study}.

\subsection{Implementation}
\label{method:impl}

\paragraph{Parallel Training with Specialized Attention Masking.}
\begin{wrapfigure}[9]{r}{0.23\textwidth}
    \centering
    \vspace{-10pt}
    \includegraphics[width=0.17\textwidth]{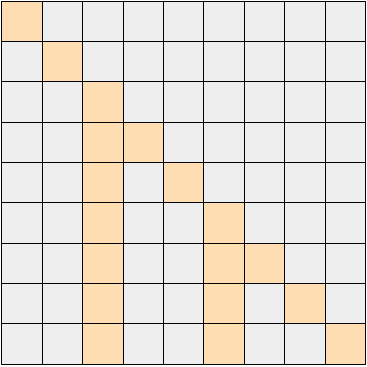}
    \caption{\textbf{Attention mask for SP training.}}
    \label{fig:sp_attn_mask}
\end{wrapfigure}
During training, scratchpad states are unrolled and concatenated into the trunk's sequence so that the loss can be computed over all byte positions in parallel,
\begin{align*}
\mathbf{z}
\;=\;
\bigl[
z_0,\,
\underbrace{z_1^1, \dots, z_1^{T_1}, z_1}_{\text{patch } 1},\,
\underbrace{z_2^1, \dots, z_2^{T_2}, z_2}_{\text{patch } 2},\,
\dots,\,
\underbrace{z_L^1, \dots, z_L^{T_L}, z_L}_{\text{patch } L}
\bigr],
\end{align*}
where $z_0$ is a \texttt{<bos>} sentinel and each patch contributes $T_\ell$ scratchpads followed by its committed representation $z_\ell$; patches with $T_\ell = 0$ collapse to $[\,\dots, z_\ell\,]$, recovering the standard patch-based layout. Self-attention in $\mathcal{M}$ is governed by a specialized causal mask (\cref{fig:sp_attn_mask}): every scratchpad or committed element of patch $\ell$ attends only to (i) itself and (ii) committed representations $z_{\ell'}$ from earlier patches $\ell' < \ell$. All scratchpads associated with patch $\ell$ share the same position index as the committed patch state. Crucially, scratchpads are never attended to by other elements, so refinement arises not from within-trunk recurrence but from the growing partial-aggregation span. This design allows all scratchpads to be processed in parallel during training and licenses their removal from the KV cache at inference.

While SP increases training-time compute by introducing transient states, the total FLOPs are comparable to training a non-SP model with a smaller patch size, e.g., one where all scratchpad positions act as patch boundaries, though the attention patterns differ due to the specialized mask.

\paragraph{Efficient Inference with Scratchpad Overriding.}
At inference time, scratchpad states are transient: only each patch's finalized representation is retained in the KV cache of the trunk $\mathcal{M}$ and exposed to subsequent patches, while scratchpads are computed on the fly and immediately overridden, incurring no additional KV-cache overhead.

\section{Experiments}
\label{experiments}

We empirically evaluate Scratchpad Patching (SP), focusing on the trade-offs among quality, persistent sequence length, and compute. We describe the experimental setup in \cref{experiments:setup}, present the main results in \cref{experiments:main}, and analyze the role of compute allocation in \cref{experiments:compute-allocation}.

\subsection{Setup}
\label{experiments:setup}
\paragraph{Models.}
All patch-based byte-level models in our experiments share the same encoder-trunk-decoder backbone (\cref{background}) and differ primarily in their patchification mechanism. We refer to each variant by its patchification strategy; labels denote the patchifier re-implemented within our shared backbone, not exact reproductions of the original model architectures, which may differ in other design choices and hyperparameters. We study four patchifier families: (i)~\textbf{Fixed-size patching} \citep{clark2022canine,nawrot2022hourglass,yu2023megabyte}, which groups bytes into non-overlapping windows of fixed width $p \in \{2, 4, 8, 16\}$; (ii)~\textbf{SpaceByte patching} \citep{slagle2024spacebyte}, which places patch boundaries at whitespace-like delimiters, producing variable-length patches; (iii)~\textbf{Entropy-based patching} \citep{nawrot2023dynamicpool,pagnoni2024blt}, where an auxiliary LM head on top of the encoder computes next-byte prediction entropy and marks positions above a threshold as patch boundaries; and (iv)~\textbf{H-Net patching} \citep{hwang2025hnet}, which uses a learned router to score each byte position and determine boundaries. For each baseline, we train and evaluate its SP variant with entropy-based scratchpad updates. We also include \textbf{byte-level} and \textbf{tokenizer-based} baselines. These main configurations have ${\sim}2$B parameters (\cref{app:model_details}), with 500M comparisons in \cref{tab:scaling}.

\paragraph{Training.}
The main configurations are pretrained on the same mixture of open-source datasets spanning code, natural language, and mathematics (\cref{app:training_data}) under a \emph{fixed-data} regime of ${\sim}400$B raw bytes. Total training FLOPs therefore differ across models, owing to their distinct average bytes per patch (or token) and scratchpad allocations. Hyperparameters are in \cref{app:training_hp}.

\paragraph{Evaluation.}
We evaluate (i) Bits-Per-Byte (BPB) on held-out validation data, (ii) estimated pass@1 on code generation with MBPP \citep{austin2021mbpp} and HumanEval \citep{chen2021humaneval}, and (iii) accuracy on multiple-choice natural language understanding benchmarks. We evaluate efficiency using the persistent \emph{sequence reduction factor} (bytes per committed trunk element) and estimated FLOPs/byte reduction relative to the byte-level baseline. Details are in \cref{app:eval_details}.

\begin{figure*}[t]
    \centering
    \includegraphics[width=\textwidth]{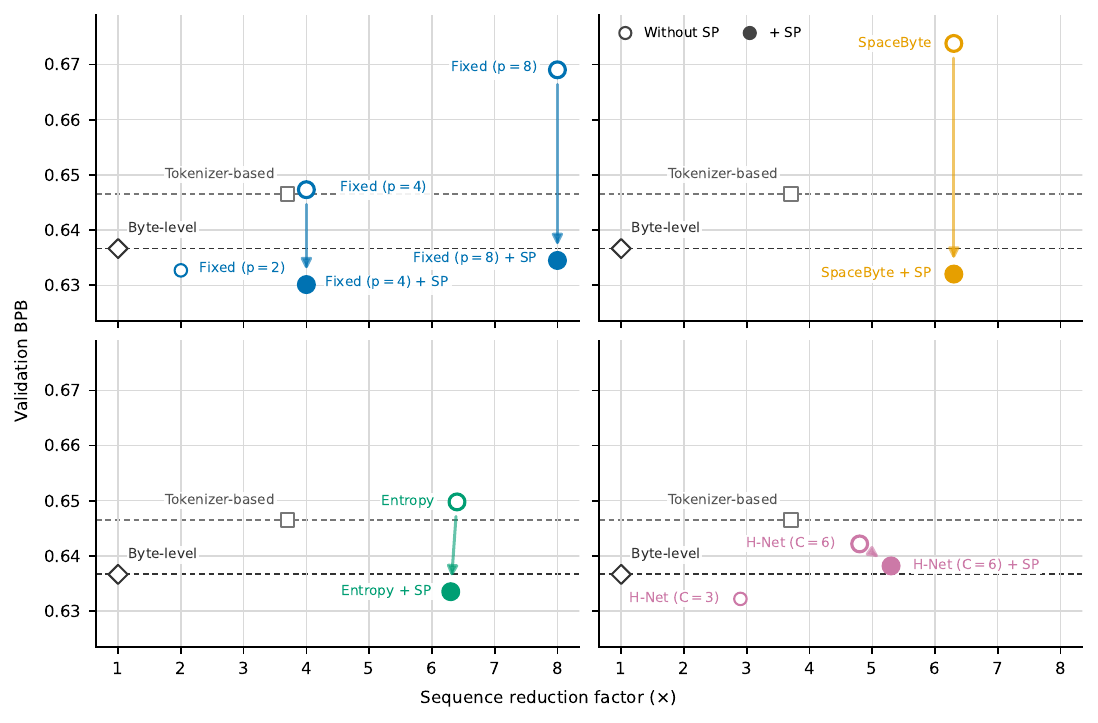}
    \caption{\textbf{Validation BPB versus sequence reduction factor, by patchifier family.} Open and filled circles denote matched non-SP and SP models; arrows connect each pair. Horizontal lines mark the byte and tokenizer baselines at $1.0\times$ and $3.7\times$ sequence reduction factors, respectively.}
    \label{fig:pareto:compression:bpb}
\end{figure*}

\subsection{Main Results}
\label{experiments:main}
\paragraph{Improved Quality-Efficiency Trade-off.}
\cref{fig:pareto:compression:bpb} plots validation BPB against the sequence reduction factor. Across all four tested patchifier families, SP lowers BPB relative to the matched non-SP model. The gains are most pronounced for larger fixed patches and SpaceByte, where vanilla models under-allocate compute to information-dense regions and suffer substantial BPB degradation. Within-patch scratchpads recover much of the quality lost to patch lag. Fixed-size and delimiter-based patching retain their committed segmentation, while learned patchifiers can change their realized patch sizes during training. Downstream comparisons are in \cref{tab:scoring,tab:code_generation,app:pareto:downstream}. Some non-SP configurations, such as Fixed~($p\!=\!2$) and H-Net~($C=3$), can outperform the byte-level baseline. SP extends these gains to larger patches, with Fixed~($p\!=\!4$)+SP and Fixed~($p\!=\!8$)+SP achieving better BPB while retaining shorter trunk sequences and FLOPs savings. The ordering is domain-dependent: Fixed~($p\!=\!8$)+SP performs better than bytes on natural language ($0.8444$ vs.\ $0.8534$) but slightly worse on code ($0.3482$ vs.\ $0.3454$), with the tokenizer showing the same split more strongly. At 500M parameters, SP still improves Fixed~($p\!=\!8$) ($0.7833 \to 0.7425$) but trails bytes ($0.7283$; \cref{tab:scaling}). We hypothesize that compression supplies useful abstractions in locally predictable regions, while entropy-triggered updates target positions most affected by patch lag.

\paragraph{Natural Language Understanding.}
\cref{tab:scoring} reports downstream accuracy on eight multiple-choice NLU benchmarks. We report sequence length and FLOPs/byte reduction measured during validation BPB evaluation as efficiency proxies. Within each patchifier family, SP improves average task accuracy and largely recovers the degradation incurred by aggressive patch sizes: Fixed~($p\!=\!16$) improves from 48.0 to 54.2 with SP, matching the byte-level baseline (54.1) despite operating at $16$ bytes per patch. After adding SP, simple schemes (e.g., fixed-size patching and SpaceByte) match or surpass more sophisticated strategies, and the gap among patchifier families narrows substantially.

The tokenizer-based model is a strong baseline on NLU tasks, outperforming the byte-level model and most non-SP patch-based models. Only Fixed~($p\!=\!4$)+SP and SpaceByte+SP marginally exceed its average of 55.9, at 56.2. We hypothesize that subword tokenization provides a useful inductive bias for language; its relative performance differs on code and across languages (\cref{analyses:multilingual}).

\begin{table*}[t]
\centering
\caption{Downstream task accuracy ($\uparrow$) on natural language understanding (NLU) benchmarks. \textbf{Bold} and \underline{underlined} entries indicate the best and second-best results per task, respectively. Sequence and FLOPs/byte reduction are measured during evaluation on validation BPB. \textsuperscript{\textdagger}SP influences the training dynamics of learned patchifiers, resulting in slightly different factors compared to the baselines.}
\label{tab:scoring}
\resizebox{\textwidth}{!}{%
\begin{tabular}{l c c c c c c c c c c c}
\toprule

\multirow{2}{*}{\textbf{Model}} & 
\textbf{Sequence} &
\textbf{FLOPs/byte} &
\multirow{2}{*}{\textbf{ARC-E}} &
\multirow{2}{*}{\textbf{ARC-C}} & 
\multirow{2}{*}{\textbf{BoolQ}} & 
\multirow{2}{*}{\textbf{HSwag}} & 
\multirow{2}{*}{\textbf{OBQA}} & 
\multirow{2}{*}{\textbf{PIQA}} &
\multirow{2}{*}{\textbf{WinoG}} &
\multirow{2}{*}{\textbf{MMLU}} &
\multirow{2}{*}{\textbf{Avg.}} \\
&
\textbf{Reduction} ($\uparrow$) &
\textbf{Reduction} ($\uparrow$) &
& & & & & & & & \\
\midrule

Byte-level                  & 1.0 & 1.0 & 66.6 & 38.7 & 64.5 & 55.1 & 45.2 & 72.8 & 56.0 & 33.5 & 54.1 \\
Tokenizer-based             & 3.7 & 4.1 & \textbf{71.6} & 40.7 & 60.7 & 58.4 & \textbf{49.2} & 73.6 & 57.8 & \textbf{35.1} & \underline{55.9} \\
\midrule

Fixed ($p\!=\!4$)               & 4.0 & 3.1 & 67.4 & 39.8 & 50.5 & 57.9 & 44.8 & 70.4 & 58.6 & 33.8 & 52.9 \\
\rowcolor{sprow} Fixed ($p\!=\!4$) + \textbf{SP} & 4.0 & 2.1 & \underline{71.0} & \underline{41.6} & 65.2 & \textbf{59.7} & 46.0 & 72.9 & \underline{58.8} & \underline{34.7} & \textbf{56.2} \\
\addlinespace
Fixed ($p\!=\!8$)               & 8.0 & 4.5 & 64.9 & 39.6 & 57.2 & 54.5 & 41.6 & 71.0 & 56.2 & 32.6 & 52.2 \\
\rowcolor{sprow} Fixed ($p\!=\!8$) + \textbf{SP} & 8.0 & 2.6 & 68.5 & 39.8 & 53.3 & 59.2 & 44.4 & 73.0 & 57.4 & 34.6 & 53.8 \\
\addlinespace
Fixed ($p\!=\!16$)              & 16.0 & 5.7 & 56.5 & 35.2 & 54.3 & 47.3 & 38.2 & 67.7 & 55.2 & 29.8 & 48.0 \\
\rowcolor{sprow} Fixed ($p\!=\!16$) + \textbf{SP}& 16.0 & 2.9 & 67.5 & 40.8 & 60.0 & 58.0 & 43.4 & 72.5 & 58.2 & 33.5 & 54.2 \\
\midrule

SpaceByte                   & 6.3 & 4.0 & 67.3 & 38.6 & 63.3 & 57.3 & 44.8 & 72.8 & 57.7 & 33.9 & 54.5 \\
\rowcolor{sprow} SpaceByte + \textbf{SP}     & 6.3 & 2.7 & 69.8 & \textbf{41.9} & 64.2 & 59.5 & \underline{48.0} & \textbf{73.8} & \textbf{59.0} & 33.7 & \textbf{56.2} \\
\midrule

Entropy-based               & 6.4 & 4.0 & 66.1 & 40.6 & 50.6 & 57.7 & 46.4 & 72.4 & 57.9 & 33.7 & 53.2 \\
\rowcolor{sprow} Entropy-based + \textbf{SP} & 6.3\textsuperscript{\textdagger} & 2.7 & 66.0 & 40.4 & \textbf{66.3} & \textbf{59.7} & 45.4 & 72.8 & 57.4 & 34.4 & 55.3 \\
\midrule

H-Net                      & 4.8 & 3.4 & 70.3 & 40.1 & 64.5 & 58.5 & 45.6 & \underline{73.7} & 56.7 & 33.9 & 55.4 \\
\rowcolor{sprow} H-Net + \textbf{SP}        & 5.3\textsuperscript{\textdagger} & 2.4 & 67.0 & 40.9 & \underline{65.9} & \underline{59.6} & 45.8 & 72.1 & \textbf{59.0} & 33.6 & 55.5 \\
\bottomrule
\end{tabular}}
\end{table*}

\paragraph{Code Generation.}
We next evaluate whether SP improves downstream \emph{generation} quality. \cref{tab:code_generation} reports pass@1 rates on MBPP and HumanEval alongside reductions in inference-time trunk KV-cache over patches and FLOPs/byte. In \cref{tab:code_generation}, SP improves pass@1 across patchifier families while largely preserving the trunk KV-cache reduction factor. At larger patch sizes, these gains come with FLOPs reduction comparable to or larger than the tokenizer baseline. Simple schemes, such as fixed-size patching and SpaceByte, are already strong baselines for code, and SP extends this advantage to large-patch regimes ($p\!=\!8$, $p\!=\!16$), recovering most of the quality lost. In contrast to the NLU setting, the tokenizer-based model is a weak baseline for code generation in our setup, underperforming both the byte-level and various patch-based models. SP-augmented models widen this gap further, while offering larger trunk KV-cache reductions than the tokenizer and preserving inference FLOPs efficiency. SP thus offers a better quality-efficiency trade-off for code generation.

\subsection{Compute Allocation Narrows the Gap Among Patchifier Choices}
\label{experiments:compute-allocation}
\begin{wrapfigure}[16]{r}{0.45\textwidth}
    \centering
    \vspace{-12pt}
    \includegraphics[width=\linewidth]{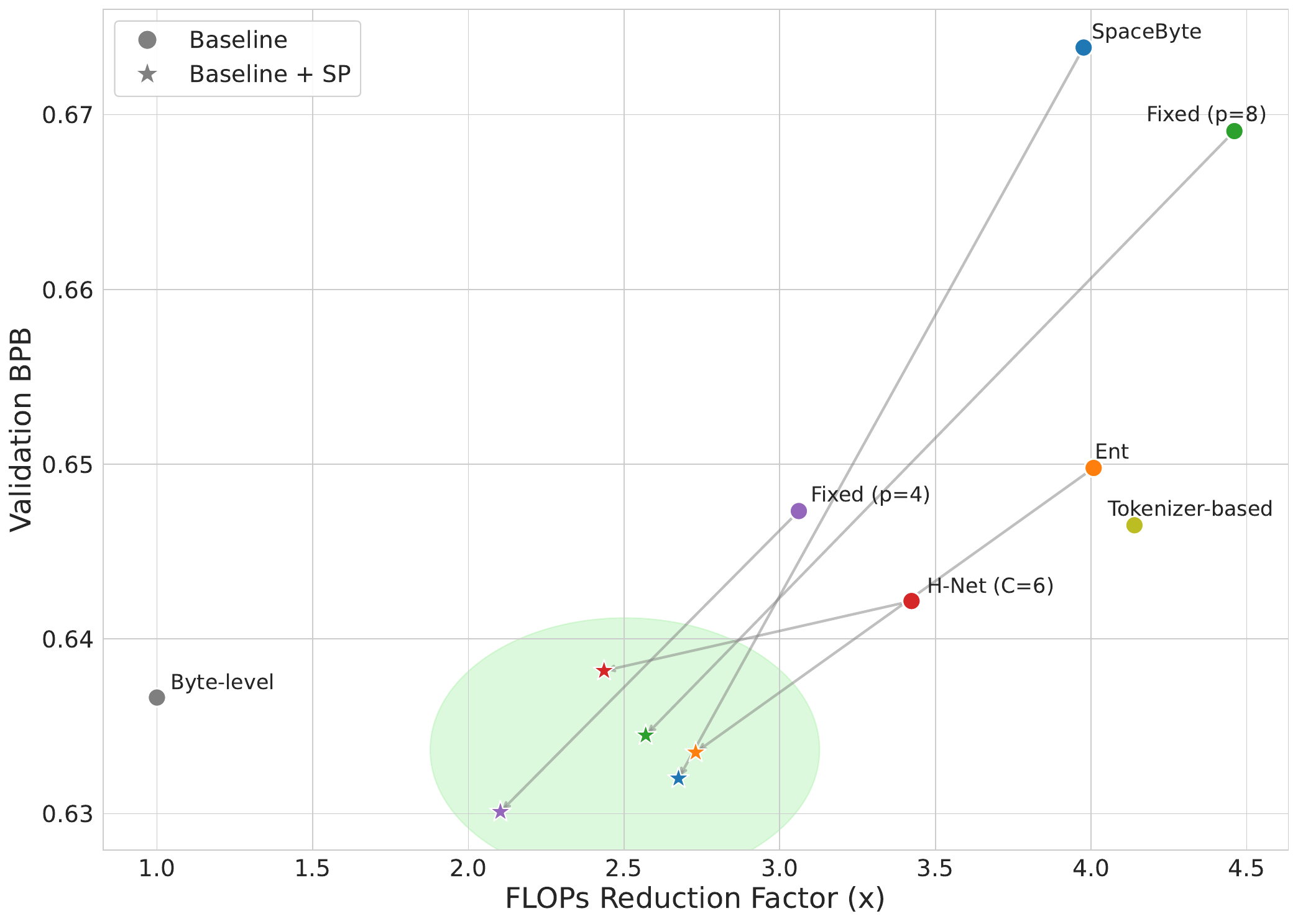}
    \caption{Validation BPB versus training FLOPs reduction relative to the byte-level baseline. The shaded region marks the cluster of SP variants.}
    \label{fig:pareto:flops:bpb}
\end{wrapfigure}
For Fixed~($p\!=\!8$), SpaceByte, entropy-based, and H\nobreakdash-Net patching, the SP/non-SP training FLOPs ratios are $1.74\times$, $1.49\times$, $1.47\times$, and $1.41\times$, respectively. These main SP variants use $2.4$--$2.7\times$ fewer training FLOPs than the byte-level baseline. At inference on the code validation set (H-Net at $\tau_\text{SP}\!=\!1.5$, as in \cref{fig:inference-time-adjustments:sp}), the estimated SP/non-SP FLOPs/byte ratios are $1.57\times$, $1.40\times$, $1.38\times$, and $1.23\times$. The results above suggest that quality differences across patchifiers are driven less by the boundary rule and more by how compute is distributed across patches. To make this explicit, \cref{fig:pareto:flops:bpb} plots validation BPB against training FLOPs reduction.

Standard patchifiers save FLOPs by shortening the trunk sequence, but at the cost of higher BPB as patch size grows. SP moves models to a better region of this trade-off: it injects additional compute via selective within-patch refinements, and does so in a way that yields disproportionately large BPB gains. After adding SP, multiple patchifier families cluster tightly in the BPB-FLOPs space, suggesting that compute allocation may matter more than the choice of patchification. We hypothesize that this is partly because attention layers within the trunk blur explicit patch boundaries, and SP compensates for imperfect or misaligned patchification by providing additional refinement opportunities.

\begin{table}[t]
\centering
\caption{Performance and efficiency comparison on MBPP and HumanEval. Trunk KV-cache and inference FLOPs/byte reductions are relative to the byte-level baseline. Trunk KV cache counts persistent patch states only. \textsuperscript{\textdagger}SP influences the training dynamics of learned patchifiers, resulting in slightly different patch sizes compared to the baselines.}
\vspace{0.5\baselineskip}
\label{tab:code_generation}
\resizebox{\textwidth}{!}{%
\begin{tabular}{l ccc ccc}
\toprule
& \multicolumn{3}{c}{\textbf{MBPP}} & \multicolumn{3}{c}{\textbf{HumanEval}} \\
\cmidrule(lr){2-4} \cmidrule(lr){5-7}
\multirow{2}{*}{\textbf{Model}}
  & \textbf{Pass@1} & \textbf{Trunk KV Cache} & \textbf{Inf. FLOPs/byte}
  & \textbf{Pass@1} & \textbf{Trunk KV Cache} & \textbf{Inf. FLOPs/byte} \\
  & (\% $\uparrow$) & \textbf{Reduction} ($\uparrow$) & \textbf{Reduction} ($\uparrow$)
  & (\% $\uparrow$) & \textbf{Reduction} ($\uparrow$) & \textbf{Reduction} ($\uparrow$) \\
\midrule
Byte-level      & 26.3 & 1.0$\times$ & 1.0$\times$ & 15.5 & 1.0$\times$ & 1.0$\times$ \\
Tokenizer-based & 23.3 & 2.6$\times$ & 3.1$\times$ & 13.3 & 3.0$\times$ & 3.5$\times$ \\
\midrule
Fixed ($p\!=\!4$)  & 28.1 & 4.0$\times$ & 3.1$\times$ & 16.8 & 4.0$\times$ & 3.1$\times$ \\
\rowcolor{sprow} Fixed ($p\!=\!4$) + \textbf{SP}  & 31.9 & 4.0$\times$ & 2.5$\times$ & 17.3 & 4.0$\times$ & 2.7$\times$ \\
\addlinespace
Fixed ($p\!=\!8$)  & 24.1 & 8.0$\times$ & 4.4$\times$ & 13.0 & 8.0$\times$ & 4.4$\times$ \\
\rowcolor{sprow} Fixed ($p\!=\!8$) + \textbf{SP}  & 32.1 & 8.0$\times$ & 3.2$\times$ & 15.9 & 8.0$\times$ & 3.7$\times$ \\
\addlinespace
Fixed ($p\!=\!16$) & 18.2 & 16.0$\times$ & 5.6$\times$ & 10.5 & 16.0$\times$ & 5.6$\times$ \\
\rowcolor{sprow} Fixed ($p\!=\!16$) + \textbf{SP} & 27.5 & 16.0$\times$ & 3.7$\times$ & 14.8 & 16.0$\times$ & 4.5$\times$ \\
\midrule
SpaceByte       & 26.5 & 5.4$\times$ & 3.7$\times$ & 15.7 & 6.3$\times$ & 4.0$\times$ \\
\rowcolor{sprow} SpaceByte + \textbf{SP}       & 30.3 & 5.4$\times$ & 2.9$\times$ & 16.6 & 6.3$\times$ & 3.5$\times$ \\
\midrule
Entropy-based   & 23.2 & 9.1$\times$ & 4.7$\times$ & 12.6 & 11.8$\times$ & 5.1$\times$ \\
\rowcolor{sprow} Entropy-based + \textbf{SP}   & 27.2 & 8.8$\times$\textsuperscript{\textdagger} & 3.3$\times$ & 16.3 & 11.7$\times$\textsuperscript{\textdagger} & 4.2$\times$ \\
\midrule
H-Net           & 27.6 & 4.7$\times$ & 3.4$\times$ & 15.1 & 5.8$\times$ & 3.8$\times$ \\
\rowcolor{sprow} H-Net + \textbf{SP}           & 28.5 & 4.7$\times$ & 3.0$\times$ & 15.4 & 6.1$\times$\textsuperscript{\textdagger} & 3.6$\times$ \\
\bottomrule
\end{tabular}}
\end{table}

\section{Analyses}
\label{analyses}

We analyze SP along several dimensions: FLOPs-matched comparisons (\cref{analyses:flops-matched-performance}), inference-time flexibility (\cref{analyses:inference-time-adjustment}), and multilingual performance (\cref{analyses:multilingual}). Additional results, including scratchpad triggering ablations and qualitative case studies, are in \cref{app:additional_results}.

\subsection{FLOPs-matched Performance Comparison}
\label{analyses:flops-matched-performance}
\begin{figure*}[t]
    \centering
    \includegraphics[width=0.9\textwidth]{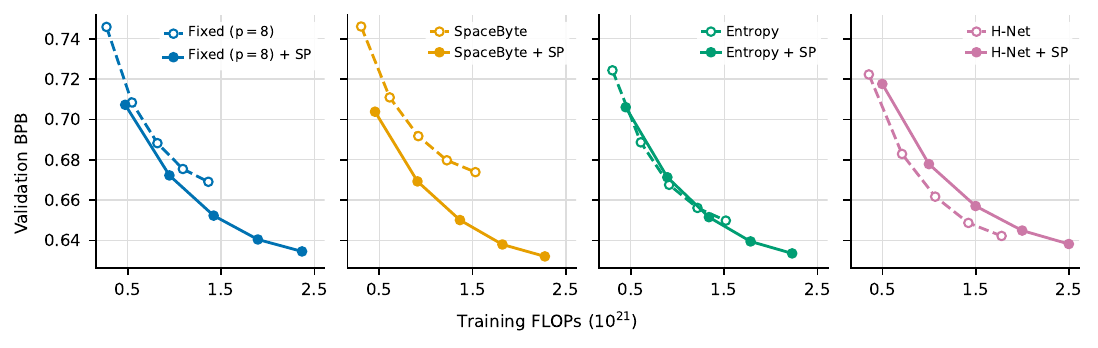}
    \vspace{-4pt}
    \caption{\textbf{Validation BPB under matched training FLOPs, broken down by different patchifier families.} Dashed and solid curves represent non-SP and SP models, respectively.}
    \label{fig:flops-matched:bpb}
    \vspace{-7pt}
\end{figure*}

We explore whether the gains from SP simply reflect additional compute by comparing SP models against their non-SP counterparts under the same total training FLOPs. As shown in \cref{fig:flops-matched:bpb}, for three of the four patchifier families, SP matches or improves BPB at equal training compute, supporting the role of better-targeted compute allocation. The main exception is H-Net: it improves from $0.6422$ to $0.6382$ BPB under fixed data but degrades under strict FLOPs matching. H-Net predicts whether the current position starts a patch, whereas the SP entropy head predicts the next byte. We hypothesize that the resulting one-position coupling can cause redundant computation (\cref{app:p_and_sp_case_study}). Dense updating also hurts natural-language BPB relative to selective triggering (\cref{analyses:sp_impl_ablation}), indicating that additional updates are not always beneficial.

\subsection{Inference-time Compute Adjustment}
\label{analyses:inference-time-adjustment}

A key practical advantage of SP is that inference-time compute can be adjusted \emph{post-hoc} without retraining. Here we vary the scratchpad update frequency at inference time; robustness to inference-time patch size variation is reported in \cref{app:inference_patch_variation}.

\begin{figure*}[t]
\centering
\begin{subfigure}[t]{0.423\textwidth}
    \centering
    \includegraphics[width=\textwidth]{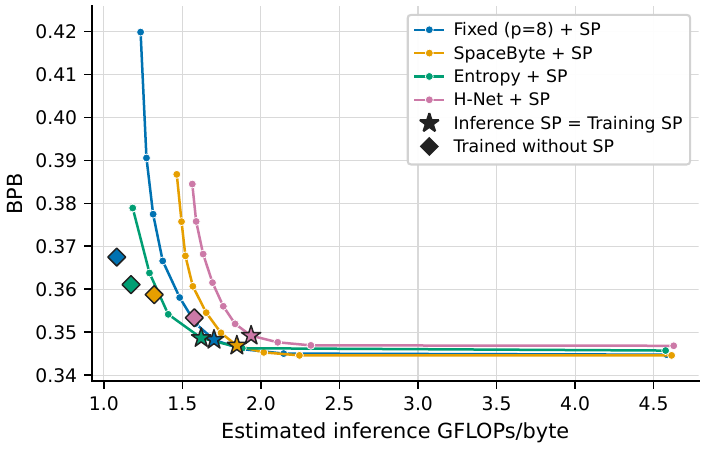}
    \caption{Validation BPB on code.}
    \label{fig:inference-time-adjustments:sp:bpb}
\end{subfigure}
\hspace{0.06\textwidth}%
\begin{subfigure}[t]{0.423\textwidth}
    \centering
    \includegraphics[width=\textwidth]{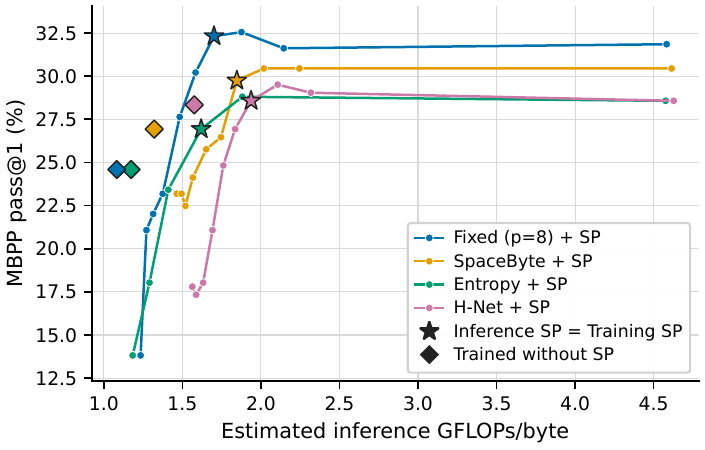}
    \caption{MBPP pass@1.}
    \label{fig:inference-time-adjustments:sp:mbpp}
\end{subfigure}
\caption{\textbf{Inference-time adjustment of scratchpad frequency.} Performance as a function of inference FLOPs/byte when varying the scratchpad update frequency post-hoc without retraining.}
\label{fig:inference-time-adjustments:sp}
\vspace{-10pt}
\end{figure*}

\cref{fig:inference-time-adjustments:sp:bpb,fig:inference-time-adjustments:sp:mbpp} show that varying the scratchpad frequency at inference time yields a generally smooth trade-off curve in both validation BPB on code and MBPP pass@1. Reducing scratchpad density relative to the training default trades compute for modest quality loss, while increasing it generally recovers performance. The entire trade-off is exposed at inference without any retraining, giving SP a single-knob compute control that standard patch-based baselines do not have.

\Needspace{15\baselineskip}
\subsection{Multilingual Performance}
\label{analyses:multilingual}

\begin{wrapfigure}[11]{r}{0.45\textwidth}
    \centering
    \vspace{-21pt}
    \includegraphics[width=0.45\textwidth]{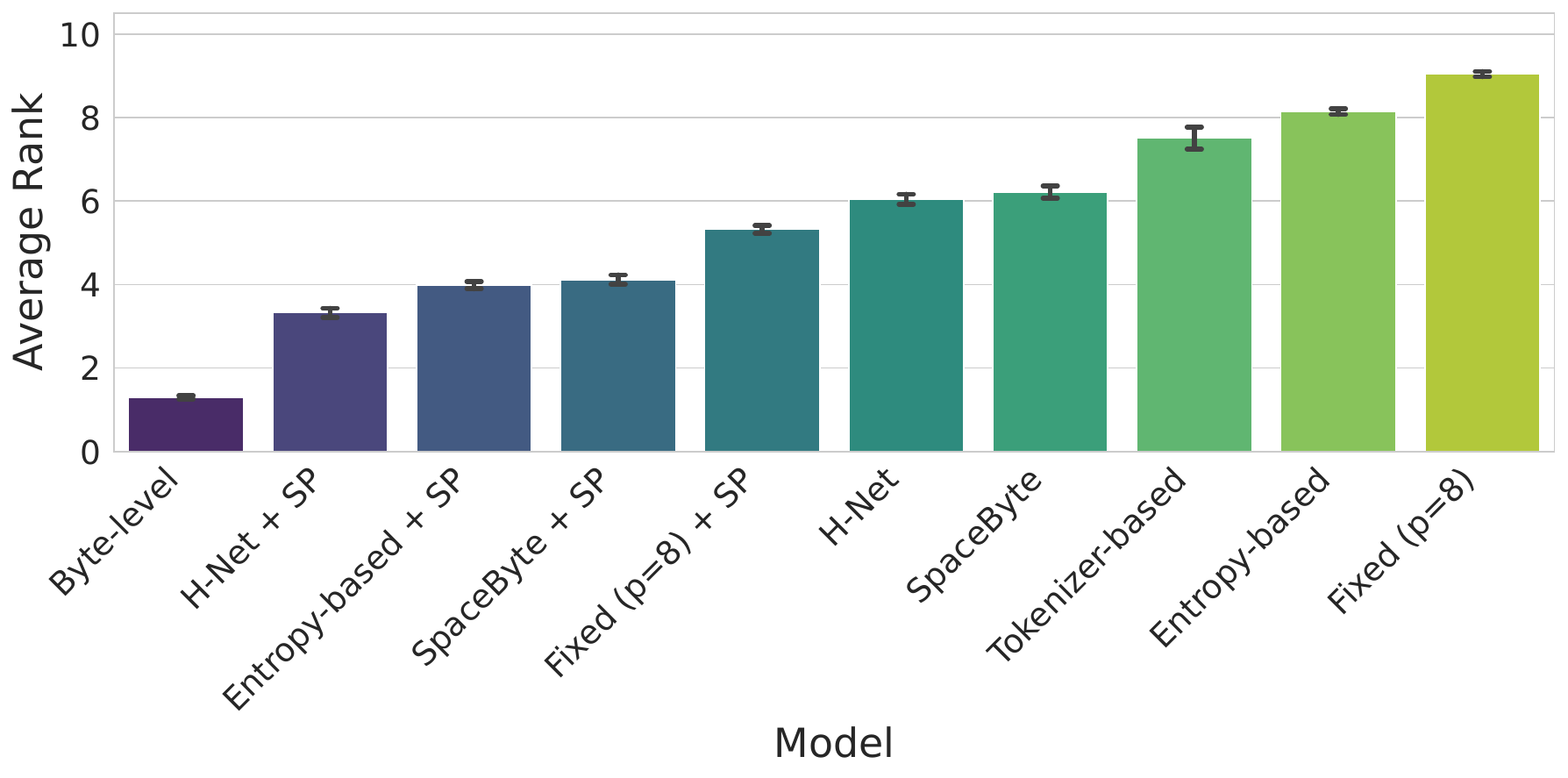}
    \caption{Average BPB rank across 200 languages of the FLORES-200 validation set. Error bars represent the standard error of the mean rank across languages.}
    \label{fig:flores_ranks}
\end{wrapfigure}

To assess robustness across languages, we evaluate all main models on the FLORES-200 validation set \citep{costa2022nllb}. \cref{fig:flores_ranks} shows the average BPB rank of each model across 200 languages (lower is better). The pure byte-level model achieves the strongest and most consistent performance, while the tokenizer-based model performs poorly on average. We hypothesize that this ordering reflects representation-specific biases tied to scripts and morphology. Adding SP consistently improves the ranking of patch-based models, narrowing the gap to the byte-level baseline across languages.

\vspace{-3pt}\section{Related Work}
\label{related_work}

\paragraph{Language Modeling Beyond Tokenization.}

The limitations of tokenizers have motivated work on alternative text representations, including morphology-driven byte encodings \citep{limisiewicz2024myte}, gzip-based and neural model-based compressors \citep{jiang2023zipclassification,lester2024training,zheng2026proxy}, concept-level semantic units \citep{barrault2024lcm}, and pixel-rendered representations \citep{salesky2021robust,rust2023pixel,lotz2023textrendering,gao2024improving}. Other work models bytes directly. ByT5 \citep{xue2022byt5} adopts T5 encoder-decoder architectures \citep{raffel2020t5} at the byte level; MrT5 \citep{kallini2024mrt5} extends ByT5 with a token deletion gating mechanism to dynamically reduce the sequence length; MambaByte \citep{wang2024mambabyte} uses Mamba \citep{gu2024mamba,dao2024mamba2,lahoti2026mamba3} layers for long byte sequences; and EvaByte \citep{zheng2025evabyte} combines multi-token prediction \citep{stern2018blockwise,gloeckle2024multitoken,cai2024medusa,grivas2025mtpc} with improved linear attention \citep{zheng22blara,zheng2023eva} for efficiency.

\vspace{-8pt}\paragraph{Patch-based Architectures for Byte Modeling.}
Patch-based byte-level models pool contiguous bytes into shorter patch-level or latent representations that are processed by a main trunk (\cref{background}). Existing designs differ primarily in how patch boundaries are determined. \emph{Static} methods group bytes into fixed-size windows, as in Funnel-Transformer \citep{dai2020funnel}, Canine \citep{clark2022canine}, Hourglass transformers \citep{nawrot2022hourglass}, MegaByte \citep{yu2023megabyte}, and Block Transformers \citep{ho2024block}, or use cross-attention to compress raw inputs to a fixed number of latent vectors, as in the Perceiver family \citep{jaegle2021perceiver,jaegle2021perceiverio,hawthorne2022perceiverar}. \emph{Dynamic} approaches adapt segmentation to the input content. Dynamic Pooling Transformers \citep[DPT;][]{nawrot2023dynamicpool} explore multiple dynamic boundary strategies spanning delimiter-based, entropy-based, and learned schemes. Delimiter-based methods place boundaries at whitespace-like positions, as also explored in SpaceByte \citep{slagle2024spacebyte} and AU-Nets \citep{videau2025aunets}. Entropy-based methods use local uncertainty to guide segmentation; Byte Latent Transformers~\citep[BLT;][]{pagnoni2024blt} leverage entropies pre-computed by a separate model offline. In contrast, the entropy-based baseline in our work applies a language modeling head on top of the encoder $\mathcal{E}$ to compute entropies online, avoiding the offline-model dependency of BLTs. Alternatively, the boundary predictor can be parameterized and learned explicitly. Charformer \citep{tay2022charformer} uses gradient-based tokenization to form latent subwords by pooling the byte sequence at multiple resolutions; MANTa \citep{godey2022manta} predicts boundaries by modeling the byte-block assignment; and H-Nets \citep{hwang2025hnet} leverage a simple yet effective design with ratio loss to enable stable boundary training, compatible with multi-stage hierarchical patching. Bolmo \citep{minixhofer2025bolmo} presents an effective framework for byteifying existing tokenizer-based models via non-causal boundary prediction aligned with tokenization behaviors. Dynamic chunking has also been applied in the image domain for spatially adaptive segmentation \citep{haridas2026dcdit}, and in DLCM \citep{qu2025dlcm} and ConceptMoE \citep{huang2026conceptmoe} for compressing tokens into higher-level concept units. Orthogonal to improving patchification schemes, our approach enables content-adaptive compute allocation that is independent of both patch size and the boundary rule, yielding a substantially better trade-off among persistent sequence length, compute, and task performance.

\vspace{-8pt}\paragraph{Adaptive Compute Allocation.}
The central idea behind SP is decoupling compute allocation from patch size or persistent sequence length, connecting to a broader line of work on adaptive computation \citep{bengio2015conditional,huang2016deep,graves2016act}. Universal transformers or looped language models \citep{dehghani2018universal,csordas2024moeut,tan2023sparse,giannou2023loopedtransformers,zhu2025ouro,wu2025plt,bae2025relaxed,geiping2025scaling,bae2025mor,zeng2026ponderlm} recurrently apply shared parameters to achieve dynamic computation, and PonderNet~\citep{banino2021pondernet} learns to control the number of recurrent steps end-to-end. Pause tokens~\citep{goyal2024pausetokens} allow models to exploit additional inference-time computation. Mixture-of-Depths~\citep{raposo2024mod} routes only a subset of tokens through each layer, while the rest bypass it via the residual connection. The PHD-Transformer \citep{wu2025phd} repeats input tokens for pretraining length scaling, retaining only the KV cache of the original tokens via a specialized attention mask; this shares with SP the principle of introducing transient states that participate in computation but are excluded from the persistent KV cache at prediction time. SP applies adaptive compute at the patch level, without increasing the trunk KV cache.

\vspace{-5pt}\section{Conclusion}
\label{conclusion}
In this work, we introduced \emph{Scratchpad Patching (SP)}, a mechanism for improving patch-based tokenizer-free byte-level models. By inserting transient scratchpad states within patches, SP decouples compute allocation from the patch sequence length and addresses \emph{patch lag}, the structural gap between next-byte predictions and the most recent patch representation available. Empirically, SP shifts the quality-efficiency frontier across four patchifier families, recovering quality lost to larger patches. Our results highlight SP as a practical step toward more flexible end-to-end language models that decide where to allocate compute, rather than inheriting it from a tokenizer or patchifier.

\vspace{-7pt}\paragraph{Limitations.}
We study single-stage patching; extending SP to multi-stage hierarchies such as H\nobreakdash-Net \citep{hwang2025hnet} or AU-Nets \citep{videau2025aunets} is left to future work. SP degrades our H-Net patchifier under matched training FLOPs (\cref{analyses:flops-matched-performance}), motivating more general scratchpad update designs that are effective for H-Net as well. In addition, we explore a simple form of scratchpads that recompute intermediate representations when new bytes become available; richer update rules that draw connections to recurrent methods may further improve efficiency and expressivity. Finally, SP increases training FLOPs by $1.41$--$1.74\times$ over non-SP patchifiers (\cref{experiments:compute-allocation}), and designing scratchpad formulations that yield more explicit compute savings remains an important direction for future research. All runs are single-seed, and efficiency is reported via analytical proxies, not measured serving latency (\cref{app:training_hp,app:eval_details}).

\renewcommand{\acksection}{\section*{Disclosure of Funding}}
\begin{ack}
\textbf{Funding.} No additional third-party funding or support was received for this work during the 36 months prior to submission.
\textbf{Competing interests.} The authors declare no additional competing financial interests beyond their listed primary affiliations.
\end{ack}

{\small
\bibliographystyle{plainnat}
\bibliography{main}
}

\newpage
\appendix
\crefalias{section}{appendix}
\crefalias{subsection}{appendix}

\section{Model Architecture Details}
\label{app:model_details}

This section describes the main 2B model architectures in our experiments. These models are designed to have roughly the same total parameter count, enabling fair comparison across different modeling paradigms. We consider three families of models: (i) a \textbf{byte-level Transformer} that operates directly on UTF-8 bytes without patching or tokenization, (ii) a \textbf{tokenizer-based Transformer} that operates on subword tokens, and (iii) \textbf{patch-based byte-level models} that group bytes into patches before processing them through a main trunk. All Transformer layers use a pre-norm design with standard multi-head attention and GEGLU feed-forward networks. \cref{tab:hyperparams} summarizes key hyperparameters for each model family.

\begin{table*}[!htb]
\centering
\caption{Model hyperparameters for the main 2B architectures. These configurations are designed to yield roughly the same total parameter count. For patch-based byte-level models, the encoder and decoder are lightweight Transformer stacks operating at byte-level resolution, while the main trunk operates on the patch-level sequence; its sequence length varies across runs depending on the realized patch size and scratchpads.}
\label{tab:hyperparams}
\begin{tabular}{l c c c c c}
\toprule
\multirow{2}{*}{\textbf{Hyperparameter}} & \multirow{2}{*}{\textbf{Byte-level}} & \multirow{2}{*}{\textbf{Tokenizer-based}} & \multicolumn{3}{c}{\textbf{Patch-based Models}} \\
\cmidrule(lr){4-6}
 &  &  & Encoder & Main Trunk & Decoder \\
\midrule
Sequence length                  & 8192   & 2216    & 8192 & ---   & 8192  \\
Hidden dim ($d_\text{model}$)    & 2048   & 2048    & 1024 & 2048  & 1024 \\
FFN hidden size ($d_\text{ff}$)  & 16384  & 16384   & 8192 & 16384 & 8192 \\
Num.\ layers ($L$)               & 18     & 16      & 4    & 16    & 4    \\
Vocab size ($|V|$)               & 320    & 100864  & 320  & ---   & ---  \\
\bottomrule
\end{tabular}
\end{table*}

\paragraph{Byte-level Transformer.}
The byte-level baseline is a standard autoregressive Transformer that predicts the next byte given all preceding bytes, without patching or tokenization. It uses a vocabulary of size 320, consisting of 256 UTF-8 byte values plus 64 reserved sentinel tokens (e.g., \texttt{<bos>} and \texttt{<pad>}). Because this embedding table and the output head are substantially smaller than those of a tokenizer-based model, we compensate by using additional Transformer layers ($18$ instead of $16$) to keep the total parameter count comparable.

\paragraph{Tokenizer-based Transformer.}
The tokenizer-based baseline follows a standard autoregressive Transformer architecture operating on subword tokens. The tokenizer has a vocabulary of size $100{,}864$ and is trained on a subsampled corpus from the training set, achieving an average of $3.7$ bytes per token and reflecting the substantial share of code data in our training mixture. The larger embedding and language modeling head account for a significant fraction of the total parameters.

\paragraph{Patch-based Byte-level Models.}
All patch-based architectures share the five-component design described in \cref{background}: an encoder~$\mathcal{E}$, a patchifier~$\mathcal{P}$, a main trunk~$\mathcal{M}$, an unpatchifier~$\mathcal{U}$, and a decoder~$\mathcal{D}$. They share the same byte-level vocabulary of size 320 as the byte-level baseline. The encoder and decoder are lightweight Transformer stacks (4 layers each, $d_\text{model} = 1024$, $d_\text{ff} = 8192$) operating at byte-level resolution, while the main trunk, which accounts for the majority of model parameters and compute, matches the dimensionality of the tokenizer-based model ($d_\text{model} = 2048$, $d_\text{ff} = 16{,}384$, $L = 16$) and processes the patch-level sequence. For the 500M patch-based models, we halve the hidden and FFN dimensions to $512/1024/512$ and $4096/8192/4096$ for the encoder/trunk/decoder, respectively, while retaining the $4/16/4$ layer counts.

We use full self-attention layers for the encoder~$\mathcal{E}$, the trunk~$\mathcal{M}$, and decoder~$\mathcal{D}$. We observed a slight regression in validation BPB when replacing them with sliding-window attention, though the difference is small. We leave more efficient encoder and decoder layer designs as future work; alternatives such as Mamba \citep{gu2024mamba,dao2024mamba2,lahoti2026mamba3} and xLSTM \citep{beck2024xlstm} are also viable and might offer a better trade-off between compute and performance, as demonstrated \citep{pagnoni2024blt,hwang2025hnet,minixhofer2025bolmo}.

These byte-level encoder and decoder layers also contribute substantially to KV-cache memory, motivating the more efficient alternatives above. The trunk KV-cache metric in \cref{tab:code_generation} tracks compression of the cached patch sequence. For intuition, assume full-width keys and values, equal precision, and the same raw-byte context. For the architectures in \cref{tab:hyperparams}, the encoder, trunk, and decoder caches have proportions $1:2:1$ for four-byte patches and $1:1:1$ for eight-byte patches. Taking the byte-level baseline as $100\%$, the combined encoder, trunk, and decoder caches occupy about $44\%$ for four-byte patches and $33\%$ for eight-byte patches. These conditional estimates exclude transient buffers and runtime overhead and are not measurements of total serving memory.

Different patch-based baselines, including fixed-size patching, SpaceByte, entropy-based patching, and H-Net models, differ only in their patchifier implementation, except for H-Net, which additionally modifies the unpatchifier. Implementation details are provided in the next section.

\section{Implementation Details}
\label{app:impl_details}
This section consolidates the implementation specifics for patch-based byte-level modeling, including patchifier and unpatchifier designs, model variants, and the SP configuration.

\subsection{Patchifier and Unpatchifier Design}
\label{app:patchifier_ablations}
The patchifier $\mathcal{P}$ partitions the byte sequence into $L$ contiguous segments $[s_\ell, e_\ell]$ for each $\ell\in\{1,\dots,L\}$. It then forms a patch-level representation $z_\ell \coloneq \operatorname{Aggregate}\left(x_{s_\ell:e_\ell}\right)$ by aggregating all byte hidden states $x_{s_\ell:e_\ell}$. In this work, we implement aggregation as a local cross-attention operation, where a summary vector $q_\ell$ queries all byte embeddings within the same patch,
\begin{align*}
    z_\ell &\coloneq \operatorname{Aggregate}\left(x_{s_\ell:e_\ell}\right) = \mathrm{CrossAttn}\!\left(q_\ell,\;x_{s_\ell:e_\ell},\;x_{s_\ell:e_\ell}\right).
\end{align*}
$q_\ell$ is calculated via mean pooling,
\begin{align*}
    q_\ell = \frac{1}{e_\ell - s_\ell + 1} \sum_{i=s_\ell}^{e_\ell} x_i.
\end{align*}
The cross-attention module is multi-headed and masked so that each patch-level summary attends only to byte positions within the corresponding patch. The resulting patch representations are then linearly projected to match the trunk dimensionality.

Symmetrically, trunk outputs pass through the unpatchifier~$\mathcal{U}$, which broadcasts them to byte positions, and are linearly projected to the decoder dimensionality and fused with the encoder residual.

We conducted ablation studies on two design choices:
\paragraph{Cross-attention.} Following \citet{pagnoni2024blt}, we investigated adding cross-attention at the patchifier (patch-level queries attending to byte-level keys/values) and at the unpatchifier (byte-level queries attending to split patch-level keys/values). We found that cross-attention at the patchifier yields noticeable BPB improvements, whereas adding it at the unpatchifier provides no measurable benefit. We therefore retain cross-attention only at the patchifier in all experiments.
 
\paragraph{Pooling Strategy.} Our default patchifier computes a mean-pooled summary over the byte-level hidden states within each patch, which then serves as the query in the cross-attention mechanism described above. We compared this against a take-last baseline \citep{hwang2025hnet} that uses the final byte-level hidden state of each patch as the query. Mean pooling yields consistent but small improvements; we adopt it as our default.

\subsection{Patch-based Model Variants}
\label{app:compressive_details}

\paragraph{Fixed-size Patching.} Following previous practices \citep{nawrot2022hourglass,yu2023megabyte}, we group bytes into non-overlapping chunks of a fixed width $p \in \{2, 4, 8, 16\}$.

\paragraph{SpaceByte Patching.} Following \citet{slagle2024spacebyte}, we use the same delimiter heuristic that ends patches at whitespace-like boundaries, producing variable-length patches with an average of roughly $6.3$ bytes per patch on our training corpus.

\paragraph{Entropy-based Patching.} We allocate two Transformer layers (with the same configuration as the encoder layer) on top of the encoder $\mathcal{E}$ output, followed by an auxiliary language modeling head that predicts the next byte. The entropy of this prediction is computed at each byte position, and positions where the entropy exceeds a threshold $\tau_\text{P}$ are marked as patch boundaries, concentrating boundaries in information-dense regions. This auxiliary head is trained jointly with the main LM head in decoder $\mathcal{D}$ by summing their losses with equal weight. We stop the gradient from the auxiliary head to the encoder to prevent it from influencing the encoder representations. We tested thresholds $\tau_\text{P} \in \{1.5, 2.5, 3\}$ and found $\tau_\text{P} = 2.5$ yields the best quality-efficiency trade-off.

\paragraph{H-Net Patching.} We closely follow the single-stage formulation of H-Net~\citep{hwang2025hnet} and implement it upon our shared architecture backbone. Similar to entropy-based patching, we add two Transformer layers on top of the encoder output. We replace the cosine-similarity scoring of the original design \citep{hwang2025hnet} with a 2-layer MLP applied to the concatenation of the current and previous byte-level hidden states, which we find to be slightly more stable and performant during training. Patch boundaries are determined by thresholding the MLP sigmoid scores at $0.5$, supervised by an auxiliary ratio loss with weight $0.03$ as in \citet{hwang2025hnet} that encourages convergence to a target average patch size $C = 6$. We also trained a variant targeting $C = 3$ for \cref{fig:pareto:compression:bpb,fig:pareto:compression:downstream}. We incorporate the smoothing operation from H-Net at the unpatchifier, which improves BPB but slows convergence to the targeted patch size; without smoothing, the model reaches the target faster but with quality degradation. We also experimented with the straight-through confidence-weighted decompression \citep{hwang2025hnet}, and found it to have marginal effects. We do not use module-wise learning rate modulations.

\subsection{Scratchpad Patching Configuration}
\label{app:sp_details}

Unless stated otherwise, scratchpad updates are triggered by an entropy-based policy: a scratchpad update is issued at byte position $n$ whenever the encoder's next-byte prediction entropy $H_n$ exceeds a threshold $\tau_\text{SP}$. We tune $\tau_\text{SP}$ over $\{0.5, 1.0, 1.5, 2.0, 2.5, 3.0\}$ per patchifier family for the best quality-efficiency trade-off, settling on $\tau_\text{SP} = 1.5$ for fixed-size and SpaceByte patching, $\tau_\text{SP} = 1.0$ for entropy-based patching, and $\tau_\text{SP} = 2.5$ for H-Net patching. Entropy-based patching admits the lower threshold because patchification and scratchpad triggering derive from the same entropy signal under $\tau_\text{P} > \tau_\text{SP}$, so scratchpads fill moderate-entropy gaps between boundaries. At most one scratchpad update is applied per byte position; extending this to multiple updates per position is a natural direction for future work. The inference-time frequency ablation in \cref{fig:inference-time-adjustments:sp} uses a separate H-Net checkpoint trained at $\tau_\text{SP}=1.5$.

\paragraph{Entropy prediction heads.}
Similar to entropy-based patching, the entropy signal is obtained from an auxiliary language modeling head built on two additional Transformer layers placed on top of the encoder, with gradients stopped at the encoder output. This auxiliary head is trained jointly with the main LM head in decoder $\mathcal{D}$ by summing their losses with equal weight.

For fixed-size and SpaceByte patching, these two layers and the output head are allocated specifically for SP. For entropy-based patching, we reuse the existing auxiliary head (already present for patch boundary decisions), maintaining two separate thresholds $\tau_\text{P} > \tau_\text{SP}$ for both patch boundaries and scratchpad updates. For H-Net, we reuse the two added Transformer layers but allocate a separate language modeling head for entropy prediction.

\paragraph{Design choices.}
We experimented with several configurations of the auxiliary entropy predictor. Stopping gradients from the auxiliary head to the encoder consistently outperforms the variant without gradient stopping. Using two Transformer layers substantially outperforms a single layer or a direct output head on top of the encoder without additional layers; however, adding more layers or increasing their capacity does not yield further improvements. We observe the same trends for the entropy-based and H-Net patch boundary predictors as well.

\section{Training Details}
\label{app:training_details}

\subsection{Training Data}
\label{app:training_data}
All main and scaling models are pretrained on a mixture of open-source datasets, consisting of DCLM \citep{li2024dclm}, Stack v2 \citep{lozhkov2024starcoder2}, and OpenWebMath \citep{paster2024openwebmath}, spanning code, natural language, and mathematics.

\subsection{Training Hyperparameters}
\label{app:training_hp}
Comparing models with different input representations involves several confounding factors: the number of unique bytes seen, total FLOPs, parameter count, effective context length, and training schedule. Within each comparison, we fix the number of raw bytes seen to ensure a \emph{fixed-data} comparison, reflecting a data-bounded training regime. Because patch-based models shorten the sequence processed by the main trunk, and SP introduces additional refinement steps, total training FLOPs vary across methods. Other optimization hyperparameters are held constant within a comparison unless otherwise specified.

The main models have roughly 2B parameters and are trained for $50{,}000$ steps with $1{,}024$ sequences of $8{,}192$ raw bytes per step on TPUs, for an approximate $400$B-byte data budget. The tokenizer-based model instead uses a context length of $2{,}216$ tokens, corresponding to roughly $8{,}192$ raw bytes at its observed compression factor. We use AdamW \citep{adamw} with a learning rate of $1 \times 10^{-3}$, $2{,}000$ warmup steps, and cosine decay to 10\% of the peak value. We set AdamW's $\epsilon = 1 \times 10^{-12}$, which we found to outperform larger values; the gain is most pronounced for models operating on byte inputs, including both the pure byte-level baseline and patch-based variants, and noticeably smaller for the tokenizer-based model. All input sequences are prepended with a \texttt{<bos>} sentinel token. Each configuration, including the main 2B / 400B-byte runs, is a single training run; we do not report seed variance.

\begin{table}[t]
\centering
\small
\caption{\textbf{Validation BPB across model scales and data budgets.} The ${\sim}100$B-byte runs use $50{,}000$ steps with sequence batch size $256$; the main runs use $1{,}024$. Each cell is one training run.}
\label{tab:scaling}
\vspace{0.5\baselineskip}
{\setlength{\tabcolsep}{10pt}%
\begin{tabular}{lccc}
\toprule
\textbf{Model} & \shortstack{\textbf{500M params.}\\\textbf{100B bytes}} & \shortstack{\textbf{2B params.}\\\textbf{100B bytes}} & \shortstack{\textbf{2B params.}\\\textbf{400B bytes}} \\
\midrule
Byte-level & 0.7283 & 0.6933 & 0.6366 \\
Tokenizer-based & 0.7509 & 0.7063 & 0.6465 \\
Fixed ($p\!=\!8$) & 0.7833 & 0.7295 & 0.6691 \\
\rowcolor{sprow} Fixed ($p\!=\!8$) + \textbf{SP} & 0.7425 & 0.6920 & 0.6345 \\
\bottomrule
\end{tabular}}
\end{table}

\section{Evaluation Details}
\label{app:eval_details}
We evaluate the main models on (i) validation Bits-Per-Byte (BPB), (ii) code generation performance, and (iii) multiple-choice natural language understanding tasks. For the 100B-byte runs at 500M and 2B parameters, we report validation BPB. Serving latency and throughput are not measured: inference FLOPs/byte and trunk KV-cache reduction are analytical proxies, and a fair wall-clock comparison would require equally optimized SP, byte-level, and tokenizer-based inference runtimes.

\paragraph{Validation BPB.} We report BPB on a held-out validation split of the training corpus. To enable finer-grained analysis, we maintain separate validation splits for code, natural language, and math data.

\paragraph{Scratchpad update frequency.}
On the validation mixture, the trunk runs once every $2.3$, $3.0$, $3.2$, $3.3$, and $2.8$ bytes on average for Fixed~($p=4$), Fixed~($p=8$), SpaceByte, entropy-based, and H-Net models with SP, respectively. These intervals count both committed patches and scratchpad updates. When a scratchpad trigger coincides with a patch boundary, only the committed patch is processed, so it counts as one trunk call.

\paragraph{Code Generation.} We evaluate on MBPP~\citep{austin2021mbpp} and HumanEval~\citep{chen2021humaneval} benchmarks, reporting pass@1 estimated from 5 samples per problem using the standard unbiased estimator of \citet{chen2021humaneval} with $n=5$ and $k=1$, under a fixed decoding configuration: temperature $0.2$ with nucleus sampling ($p = 0.95$). We use 3-shot prompting with the standard prompts from MBPP and zero-shot prompting for HumanEval.

\paragraph{Natural Language Understanding.} We evaluate on a suite of eight multiple-choice downstream benchmarks spanning commonsense reasoning, reading comprehension, and broad knowledge: \textbf{ARC-Easy and ARC-Challenge}~\citep{clark2018arc}, \textbf{BoolQ}~\citep{clark2019boolq}, \textbf{HellaSwag}~\citep{zellers2019hellaswag}, \textbf{OpenBookQA (OBQA)}~\citep{mihaylov2018openbookqa}, \textbf{PIQA}~\citep{bisk2020piqa}, \textbf{WinoGrande}~\citep{sakaguchi2020winograde}, and \textbf{MMLU}~\citep{hendrycks2021mmlu}. For all tasks, we follow the OLMES evaluation protocol~\citep{gu2024olmes} with curated 5-shot prompting.

\section{Additional Experimental Results}
\label{app:additional_results}

\subsection{Downstream Quality-Efficiency Pareto Frontiers}
\label{app:pareto:downstream}

\begin{figure*}[t]
\centering
\includegraphics[width=\textwidth]{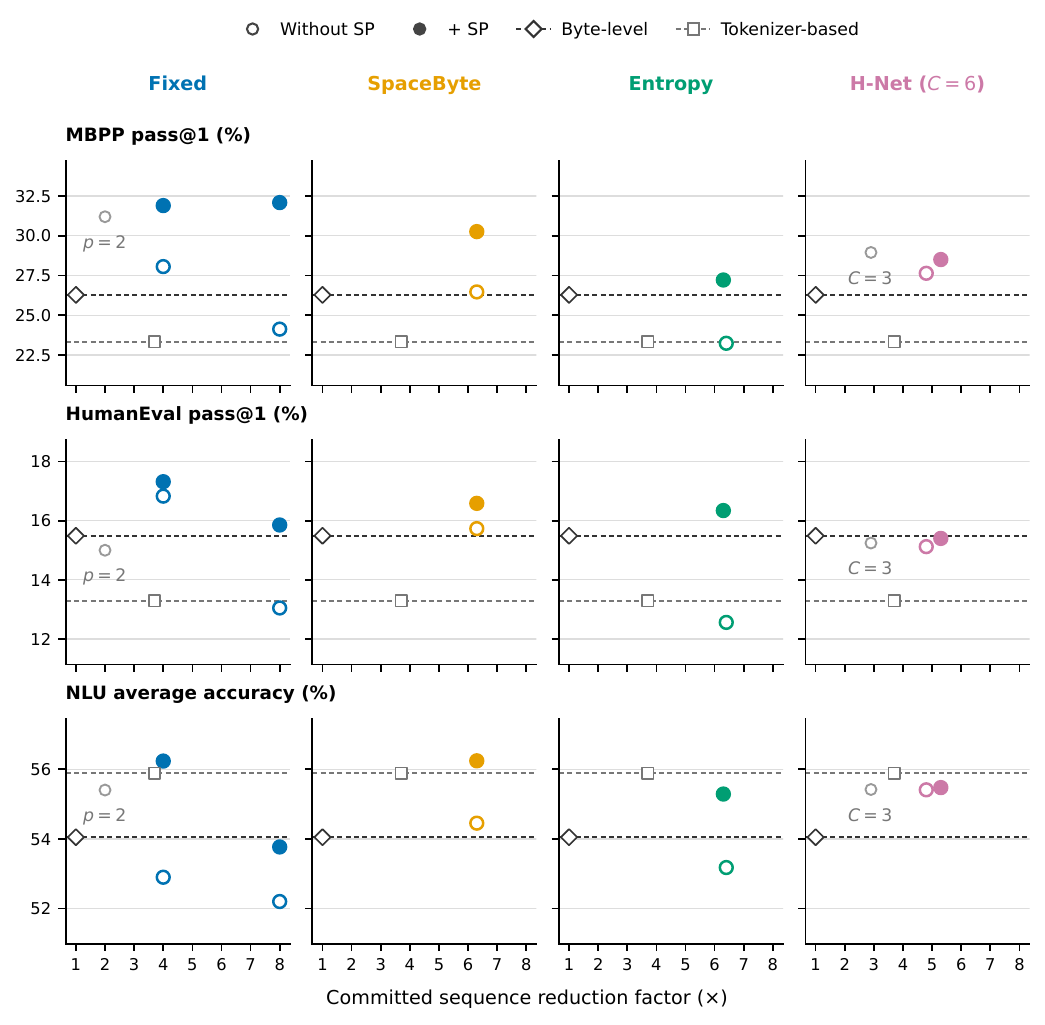}
\caption{\textbf{Downstream performance versus committed sequence reduction, by metric (rows) and patchifier family (columns).} Open and filled circles denote non-SP and SP models; gray circles show the labeled additional non-SP configurations. Diamonds and squares identify the byte-level and tokenizer baselines, with horizontal reference lines. Each row shares a y-axis scale. The sequence reduction factor is measured under the training configuration; for inference-time FLOPs and KV-cache reduction on downstream tasks, see \cref{tab:scoring,tab:code_generation}.}
\label{fig:pareto:compression:downstream}
\end{figure*}

\cref{fig:pareto:compression:downstream} shows that SP improves every matched pair on MBPP, HumanEval, and average NLU. The largest recovery occurs for aggressive fixed patching, consistent with the BPB results in \cref{fig:pareto:compression:bpb}. The H-Net generation results agree with \cref{tab:code_generation}.

\subsection{Ablations of Scratchpad Patching Strategies}
\label{analyses:sp_impl_ablation}

\begin{figure*}[ht]
\centering
\begin{subfigure}[t]{0.32\textwidth}
    \centering
    \includegraphics[width=\textwidth]{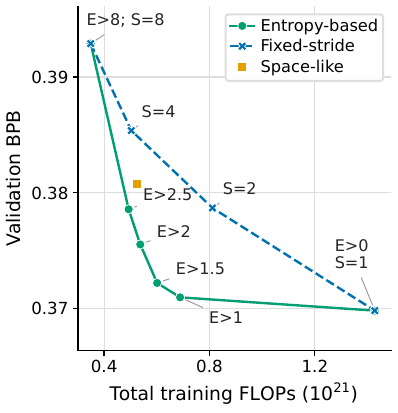}
    \caption{Validation BPB on code.}
    \label{fig:sp_ablation:code}
\end{subfigure}
\hfill
\begin{subfigure}[t]{0.32\textwidth}
    \centering
    \includegraphics[width=\textwidth]{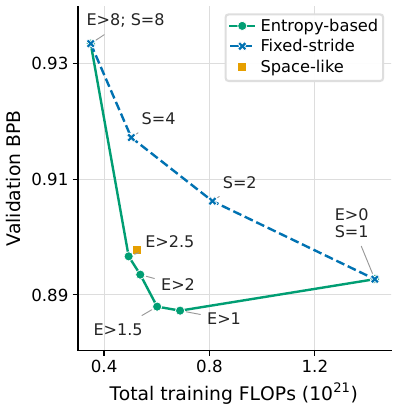}
    \captionsetup{justification=centering}
    \caption{Validation BPB on\\natural language.}
    \label{fig:sp_ablation:nl}
\end{subfigure}
\hfill
\begin{subfigure}[t]{0.32\textwidth}
    \centering
    \includegraphics[width=\textwidth]{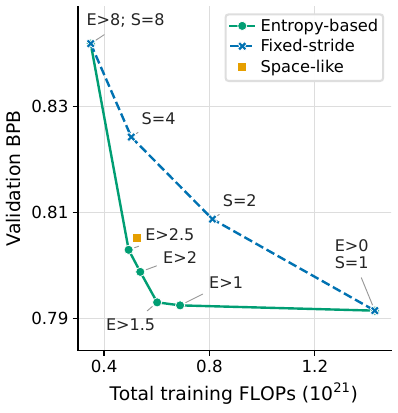}
    \caption{Validation BPB on math data.}
    \label{fig:sp_ablation:math}
\end{subfigure}
\caption{\textbf{Ablations of scratchpad triggering strategies} on validation BPB versus estimated total training FLOPs, using fixed-size patching ($p\!=\!8$) and a ${\sim}100$B-byte training budget. Entropy-based triggers ($E > \tau_\text{SP}$), fixed-stride updates ($S$), and whitespace-based heuristics are compared. The top-left point ($\tau_\text{SP} = 8$ or $S = 8$) corresponds to the non-SP baseline; the rightmost point ($\tau_\text{SP} = 0$ or $S = 1$) applies dense byte-level compute while \emph{retaining the same committed patch sequence}.}
\label{fig:sp_ablation}
\end{figure*}

We ablate the triggering strategy used for scratchpad updates, using fixed-size patching with patch size $p\!=\!8$ as the base patchifier. Our default policy issues a scratchpad update whenever the encoder's next-byte prediction entropy exceeds a threshold, where we use $\tau_\text{SP} = 1.5$ by default. We compare against alternative strategies: entropy thresholds (denoted as $E > \tau_\text{SP}$) with $\tau_\text{SP} \in \{0, 1, 1.5, 2, 2.5, 8\}$, fixed-stride updates every $S \in \{1, 2, 4, 8\}$ positions, and updates on whitespace-like bytes. The two extremes are instructive: $\tau_\text{SP} = 8$ or stride $S = 8$ effectively suppresses all scratchpads, recovering the non-SP baseline; $\tau_\text{SP} = 0$ or $S = 1$ calls the trunk at every byte position, using scratchpads except at committed patch boundaries. Importantly, this setting is \emph{not} equivalent to a standard byte-level baseline: the model still retains the same number of persistent patches for the KV cache, but allocates byte-level compute via transient scratchpad states. This highlights a unique advantage of SP, which decouples compute allocation from the committed patch sequence, enabling byte-level compute with patch-level memory efficiency.

\cref{fig:sp_ablation} reveals several findings. First, entropy-based triggering generally achieves the best compute-quality trade-off across all three domains, and all strategies, including simple whitespace-based heuristics, improve consistently over the non-SP baseline, confirming that within-patch scratchpads are beneficial regardless of the triggering policy.

Second, the three domains exhibit a revealing difference in how they respond to dense updates. On code and math (\cref{fig:sp_ablation:code,fig:sp_ablation:math}), the gap between $\tau_\text{SP} = 1.0$ and $\tau_\text{SP} = 0$ (every-position updates) is marginal, despite the latter requiring substantially more training FLOPs. This indicates that selective, entropy-guided scheduling captures most of the benefit of dense refinement at a fraction of the cost, and further compute yields diminishing returns. On natural language (\cref{fig:sp_ablation:nl}), however, dense updates yield slightly \emph{worse} BPB than selective thresholds such as $\tau_\text{SP} = 1.0$ or $1.5$. We hypothesize that this is related to the entropy profile of natural language: the majority of byte positions in prose are highly predictable (e.g., completing common words), and forcing the model to re-process these trivial positions can dilute the training signal and cause the model to overfit to local patterns at the expense of long-range dependencies. Code and math may offer fewer such predictable positions (e.g., syntax, operators, variable names, and numerical expressions), so dense updates may waste less capacity on trivial bytes and continue to yield modest gains. These results suggest that selective, entropy-guided scheduling is not only more efficient but can be actively preferable to dense refinement.

\subsection{Inference-time Patch Size Variation}
\label{app:inference_patch_variation}
\begin{figure*}[t]
\centering
\begin{subfigure}[t]{0.47\textwidth}
    \centering
    \patchSizePanel{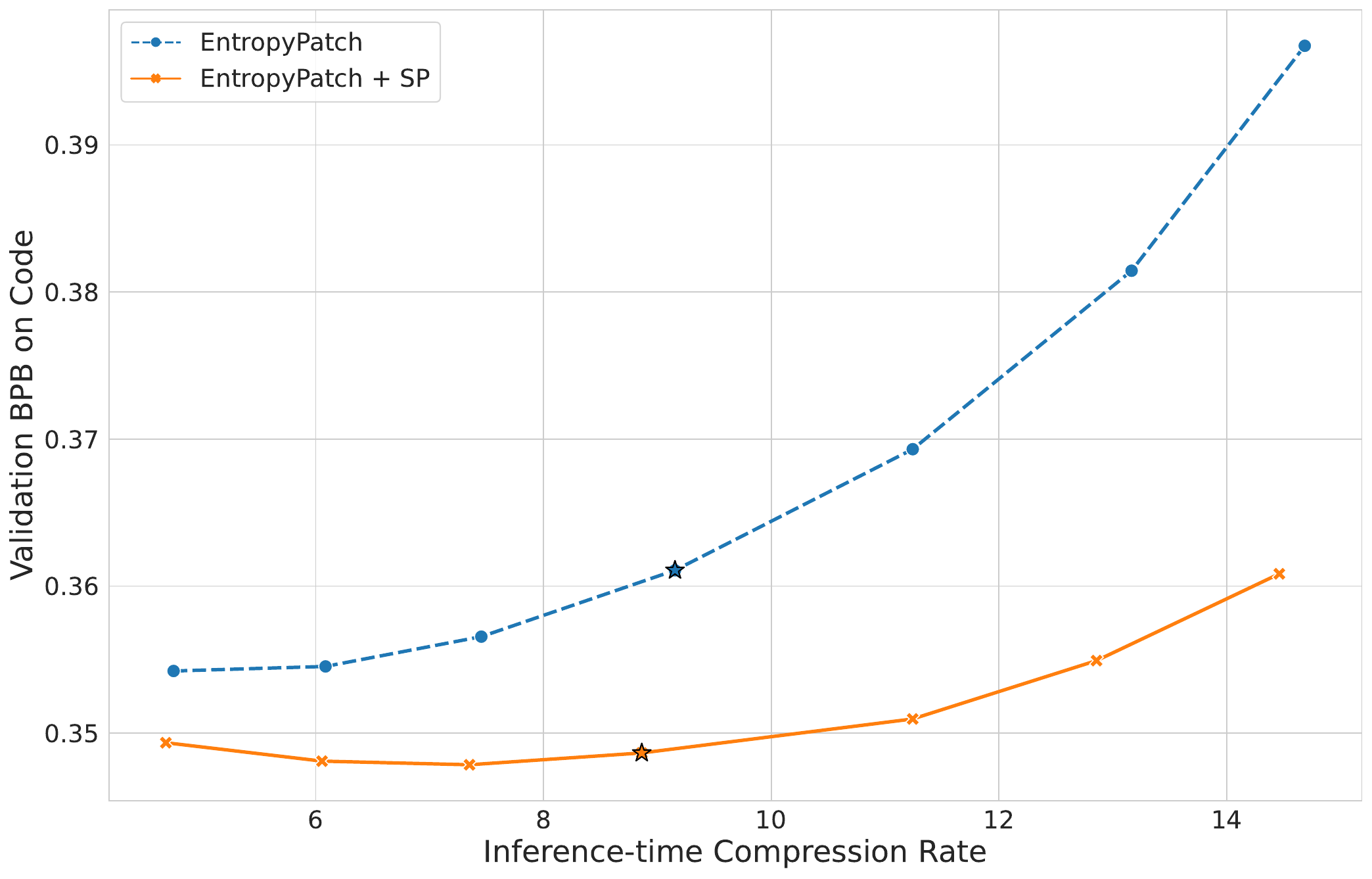}
    \caption{Validation BPB on code.}
    \label{fig:inference-time-adjustments:patch:bpb}
\end{subfigure}
\hfill
\begin{subfigure}[t]{0.47\textwidth}
    \centering
    \patchSizePanel{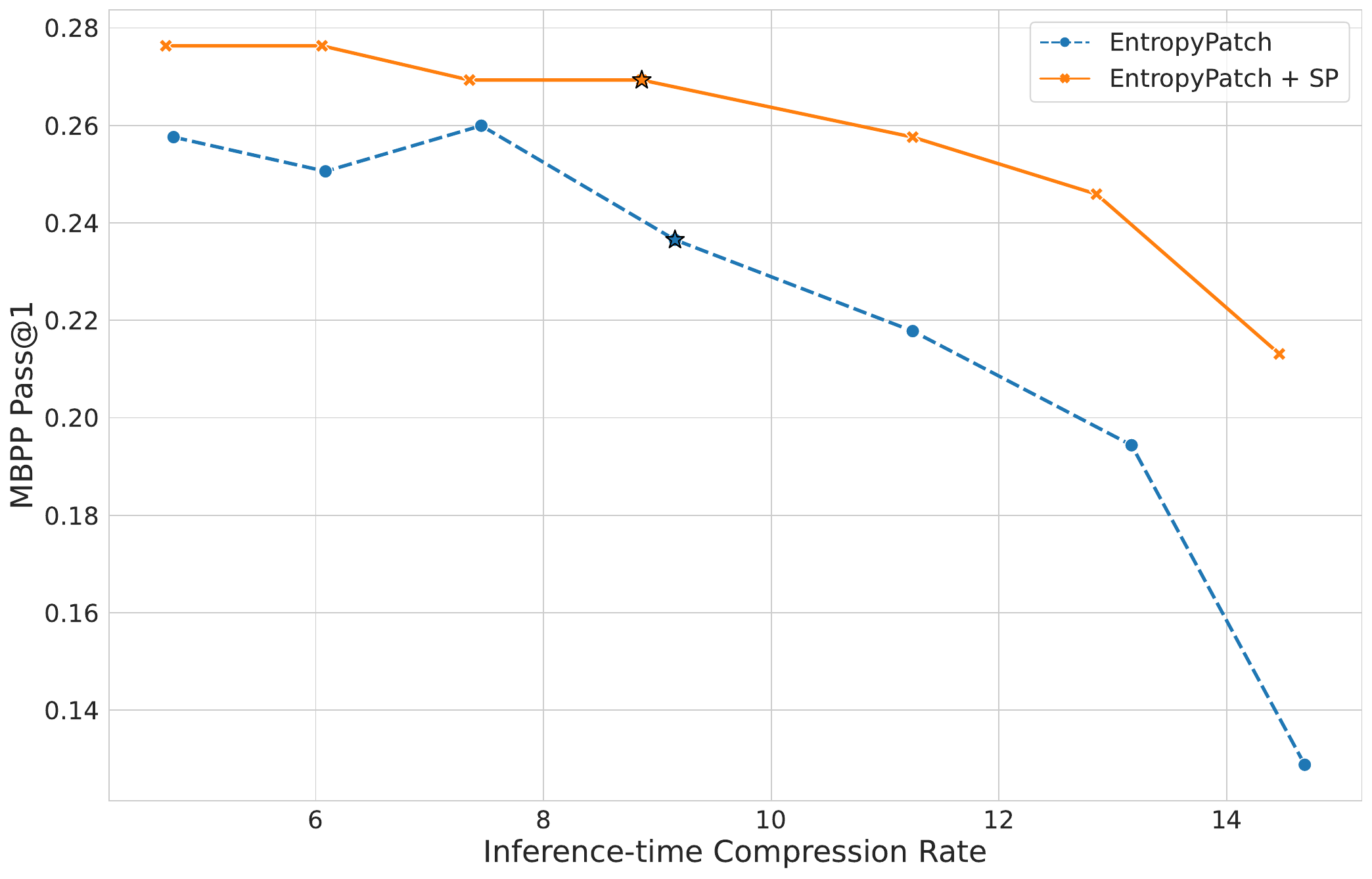}
    \caption{MBPP pass@1.}
    \label{fig:inference-time-adjustments:patch:mbpp}
\end{subfigure}
\caption{\textbf{Inference-time patch size variation.} Performance under different realized average patch sizes applied at inference time without retraining.}
\label{fig:inference-time-adjustments:patch}
\end{figure*}

Complementary to scratchpad frequency (\cref{analyses:inference-time-adjustment}), we also vary the patch size at inference time. We use entropy-based patching with threshold $\tau_{\text{P}} = 2.5$ as the default model and vary $\tau_{\text{P}}$ at inference time to induce different realized average patch sizes. As shown in \cref{fig:inference-time-adjustments:patch:bpb,fig:inference-time-adjustments:patch:mbpp}, models trained with SP exhibit strong robustness to inference-time changes in patch size: performance degrades gracefully when the realized patch size deviates from the training configuration. In contrast, non-SP models suffer substantial performance drops under the same mismatch, suggesting that SP can substantially compensate for suboptimal patch boundaries.

\subsection{Patchification Behavior}
\label{app:p_and_sp_case_study}

\begin{figure*}[thbp]
    \centering
    \includegraphics[width=\textwidth]{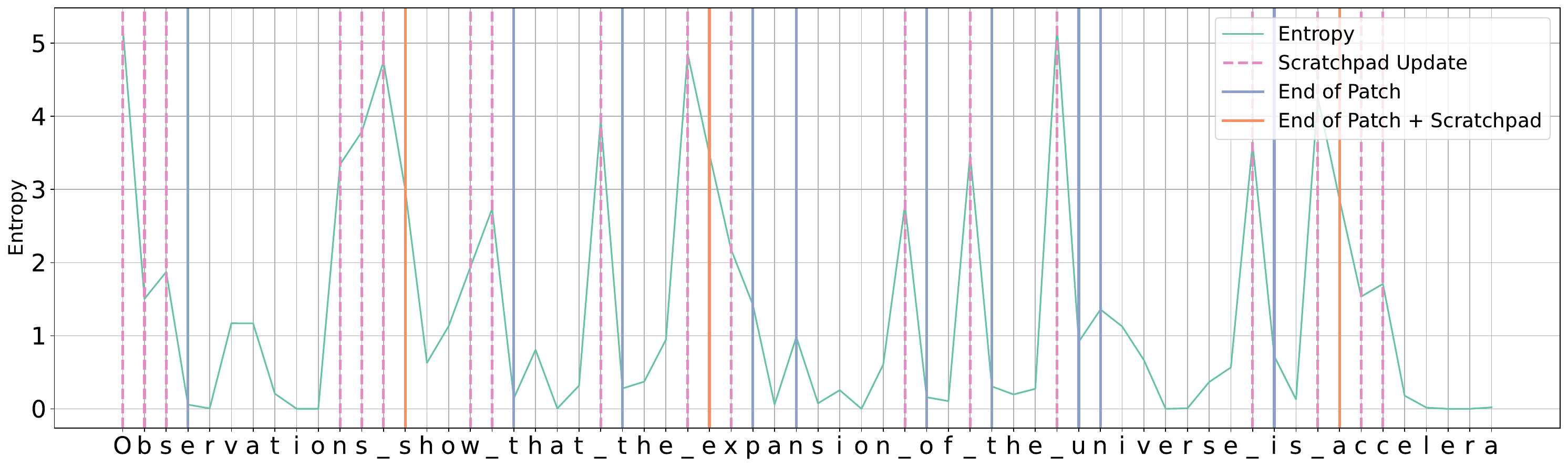} 
    \caption{\textbf{Scratchpad Patching dynamics on H-Net patching.} Patch boundaries (\textbf{solid blue}) are determined by the learned score, while scratchpad updates (\textbf{dashed pink}) fire when the encoder's next-byte entropy (\textbf{green}) exceeds the threshold. When a scratchpad-trigger coincides with a patch boundary, patchification takes precedence (\textbf{solid orange}). We use $\tau_\text{SP} = 1.5$ in this demo (main results use $\tau_\text{SP} = 2.5$; \cref{app:sp_details}) to highlight the strong spatial coupling between scratchpad triggers and patch boundaries: scratchpad triggers frequently fall one position before a patch boundary, which may produce redundant compute on locally correlated adjacent bytes.}
    \label{fig:patching_demo:hnet:text}
\end{figure*}

While the main text illustrates SP on fixed-size patching (\cref{fig:patching_demo:fixed8:text}), here we qualitatively examine how SP interacts with the other three patchifier families. The H-Net patchifier (\cref{fig:patching_demo:hnet:text}) exhibits strong spatial coupling between scratchpad updates and patch boundaries, with updates often offset by a single position. We hypothesize that this coupling causes redundant computation on locally correlated adjacent bytes, helping explain why SP yields smaller gains (and occasionally inefficiencies) for such patchifiers.

\begin{figure*}[t]
    \centering
    \includegraphics[width=\textwidth]{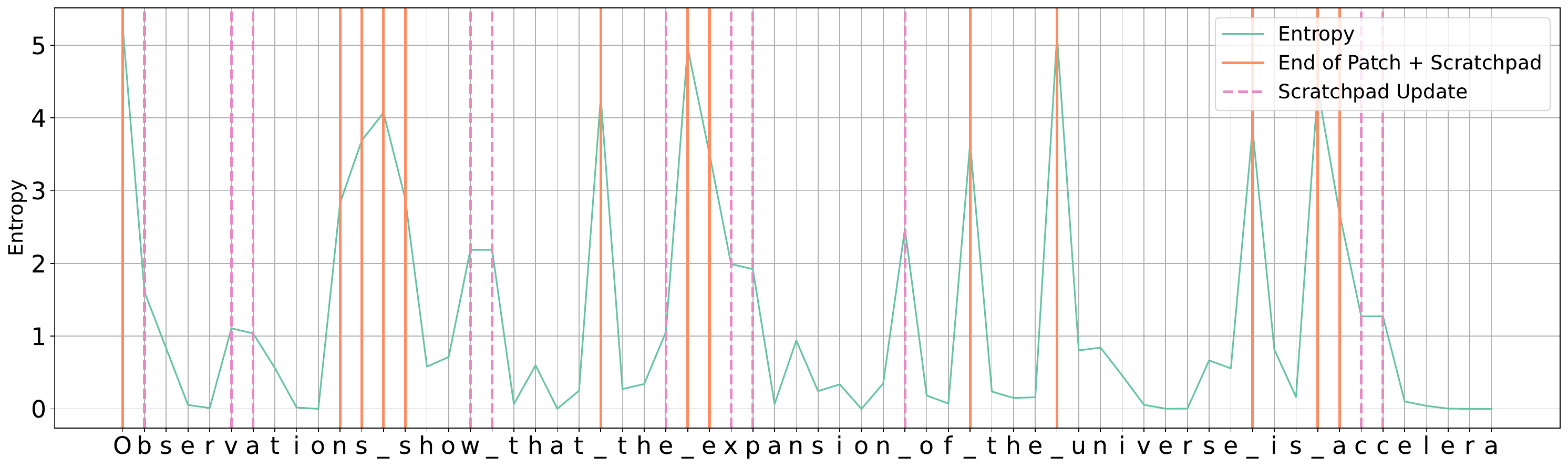} 
    \caption{\textbf{Scratchpad Patching dynamics on entropy-based patching.} Patch boundaries (\textbf{solid blue}) are placed where the encoder's next-byte entropy (\textbf{green}) exceeds the patching threshold $\tau_\text{P} = 2.5$, while scratchpad updates (\textbf{dashed pink}) fire with the lower threshold $\tau_\text{SP} = 1.0$. When a scratchpad-trigger coincides with a patch boundary, patchification takes precedence (\textbf{solid orange}).}
    \label{fig:patching_demo:entropy:text}
\end{figure*}

\begin{figure*}[t]
    \centering
    \includegraphics[width=\textwidth]{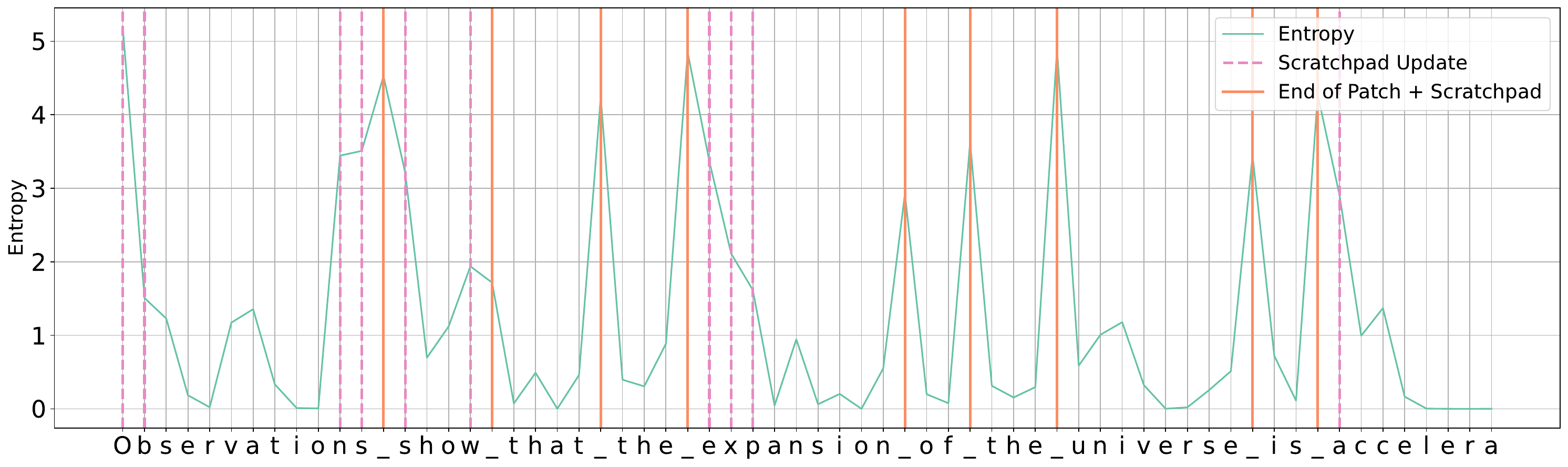} 
    \caption{\textbf{Scratchpad Patching dynamics on SpaceByte patching.} Patch boundaries (\textbf{solid blue}) are placed at whitespace-like delimiters, while scratchpad updates (\textbf{dashed pink}) fire when the encoder's next-byte entropy (\textbf{green}) exceeds threshold $\tau_\text{SP} = 1.5$. When a scratchpad-trigger coincides with a patch boundary, patchification takes precedence (\textbf{solid orange}).}
    \label{fig:patching_demo:spacebyte:text}
\end{figure*}

SP also integrates naturally with other patchifier families. For SpaceByte (\cref{fig:patching_demo:spacebyte:text}), scratchpad updates frequently coincide with patch boundaries; since patchification takes precedence in such cases (\cref{method:main}), no redundant computation is introduced. For entropy-based patching (\cref{fig:patching_demo:entropy:text}), scratchpads and patch boundaries are coordinated through distinct thresholds, separating scratchpad updates from committed boundary decisions. These qualitative patterns are consistent with the quantitative results in earlier sections and underscore that compatibility between patchification and scratchpad scheduling is critical for effective compute use.

\clearpage
\section*{NeurIPS Paper Checklist}

\begin{enumerate}

\item {\bf Claims}
    \item[] Question: Do the main claims made in the abstract and introduction accurately reflect the paper's contributions and scope?
    \item[] Answer: \answerYes{} %
    \item[] Justification: The abstract and introduction state the core contributions: introducing Scratchpad Patching to reduce patch lag, improving the quality-efficiency frontier across the four tested patchifier families, and narrowing the performance gap among patchifier choices under comparable compute. These claims are supported by experiments in \cref{experiments:main,experiments:compute-allocation} and analyses in \cref{analyses:flops-matched-performance}.

\item {\bf Limitations}
    \item[] Question: Does the paper discuss the limitations of the work performed by the authors?
    \item[] Answer: \answerYes{} %
    \item[] Justification: The Limitations paragraph in \cref{conclusion} discusses the focus on single-stage patching, the simple scratchpad update rule, the H-Net degradation under matched FLOPs, the $1.41$--$1.74\times$ training-FLOPs increase over non-SP patchifiers, single-run experiments, and the absence of measured serving latency (details in \cref{app:training_hp,app:eval_details}).

\item {\bf Theory assumptions and proofs}
    \item[] Question: For each theoretical result, does the paper provide the full set of assumptions and a complete (and correct) proof?
    \item[] Answer: \answerNA{} %
    \item[] Justification: The paper makes empirical contributions and does not include formal theoretical results.

\item {\bf Experimental result reproducibility}
    \item[] Question: Does the paper fully disclose all the information needed to reproduce the main experimental results of the paper to the extent that it affects the main claims and/or conclusions of the paper (regardless of whether the code and data are provided or not)?
    \item[] Answer: \answerYes{}
    \item[] Justification: The paper specifies the model architectures in \cref{app:model_details}, patchifier and SP implementations in \cref{app:impl_details}, training setup and hyperparameters in \cref{app:training_details,app:training_hp}, and evaluation protocols in \cref{app:eval_details}.

\item {\bf Open access to data and code}
    \item[] Question: Does the paper provide open access to the data and code, with sufficient instructions to faithfully reproduce the main experimental results, as described in supplemental material?
    \item[] Answer: \answerNo{} %
    \item[] Justification: We do not release training code, model checkpoints, or generated model artifacts, due to constraints associated with the research setting and implementation infrastructure. The training corpora and evaluation benchmarks are publicly available, and the paper provides architectural, training, and evaluation details to support reproduction.

\item {\bf Experimental setting/details}
    \item[] Question: Does the paper specify all the training and test details (e.g., data splits, hyperparameters, how they were chosen, type of optimizer) necessary to understand the results?
    \item[] Answer: \answerYes{}
    \item[] Justification: Training data, model scale, context lengths, optimizer, learning-rate schedule, batch size, and training duration are described in \cref{app:training_data,app:training_hp}; patchifier and SP hyperparameters are described in \cref{app:compressive_details,app:sp_details}; and evaluation settings, decoding strategies, and benchmark protocols are described in \cref{app:eval_details}.

\item {\bf Experiment statistical significance}
    \item[] Question: Does the paper report error bars suitably and correctly defined or other appropriate information about the statistical significance of the experiments?
    \item[] Answer: \answerNo{} %
    \item[] Justification: Pretraining $\sim$2B-parameter models multiple times per configuration is computationally prohibitive, so we do not report error bars on the BPB or NLU results. To reduce variance on code generation, pass@1 is estimated from 5 samples per problem using the unbiased estimator of \citet{chen2021humaneval}; \cref{fig:flores_ranks} reports the standard error of the mean rank across the 200 FLORES languages.

\item {\bf Experiments compute resources}
    \item[] Question: For each experiment, does the paper provide sufficient information on the computer resources (type of compute workers, memory, time of execution) needed to reproduce the experiments?
    \item[] Answer: \answerYes{}
    \item[] Justification: We report model sizes, training steps, data budget, and context lengths in \cref{app:model_details,app:training_hp}, as well as estimated training FLOPs in \cref{analyses:flops-matched-performance} as a hardware-agnostic measure of compute. Experiments were conducted on TPUs; each main run trains a $\sim$2B-parameter model for 50{,}000 steps on $\sim$400B bytes. The full research project consumed substantially more compute than the reported experiments, due to preliminary exploration and ablations not included in the paper.

\item {\bf Code of ethics}
    \item[] Question: Does the research conducted in the paper conform, in every respect, with the NeurIPS Code of Ethics \url{https://neurips.cc/public/EthicsGuidelines}?
    \item[] Answer: \answerYes{} %
    \item[] Justification: The research conforms to the NeurIPS Code of Ethics.

\item {\bf Broader impacts}
    \item[] Question: Does the paper discuss both potential positive societal impacts and negative societal impacts of the work performed?
    \item[] Answer: \answerNA{} %
    \item[] Justification: This work is a foundational methodology for patch-based byte-level language modeling and is not tied to a specific deployment or downstream application. Potential societal impacts are indirect and inherited from general improvements in language model efficiency and capability rather than being specific to the proposed mechanism.
    
\item {\bf Safeguards}
    \item[] Question: Does the paper describe safeguards that have been put in place for responsible release of data or models that have a high risk for misuse (e.g., pre-trained language models, image generators, or scraped datasets)?
    \item[] Answer: \answerNA{} %
    \item[] Justification: The paper does not release pretrained models, datasets, or other assets that pose dual-use or misuse risks.

\item {\bf Licenses for existing assets}
    \item[] Question: Are the creators or original owners of assets (e.g., code, data, models), used in the paper, properly credited and are the license and terms of use explicitly mentioned and properly respected?
    \item[] Answer: \answerYes{}
    \item[] Justification: We use publicly available training corpora and evaluation benchmarks, each cited in \cref{app:training_data} and \cref{app:eval_details} and used in accordance with their respective licenses.

\item {\bf New assets}
    \item[] Question: Are new assets introduced in the paper well documented and is the documentation provided alongside the assets?
    \item[] Answer: \answerNA{} %
    \item[] Justification: The paper does not release new datasets, model checkpoints, or code artifacts.

\item {\bf Crowdsourcing and research with human subjects}
    \item[] Question: For crowdsourcing experiments and research with human subjects, does the paper include the full text of instructions given to participants and screenshots, if applicable, as well as details about compensation (if any)? 
    \item[] Answer: \answerNA{} %
    \item[] Justification: The paper does not involve crowdsourcing or research with human subjects.

\item {\bf Institutional review board (IRB) approvals or equivalent for research with human subjects}
    \item[] Question: Does the paper describe potential risks incurred by study participants, whether such risks were disclosed to the subjects, and whether Institutional Review Board (IRB) approvals (or an equivalent approval/review based on the requirements of your country or institution) were obtained?
    \item[] Answer: \answerNA{} %
    \item[] Justification: The paper does not involve research with human subjects.

\item {\bf Declaration of LLM usage}
    \item[] Question: Does the paper describe the usage of LLMs if it is an important, original, or non-standard component of the core methods in this research? Note that if the LLM is used only for writing, editing, or formatting purposes and does \emph{not} impact the core methodology, scientific rigor, or originality of the research, declaration is not required.
    \item[] Answer: \answerNA{} %
    \item[] Justification: LLMs were used only for proofreading and minor writing polish, and did not affect the core methodology, experimental design, or scientific contributions.

\end{enumerate}

\end{document}